\documentclass[runningheads]{llncs}

\usepackage{eccv}

\usepackage{eccvabbrv}

\usepackage{graphicx}
\usepackage{booktabs}

\usepackage[accsupp]{axessibility}  

\usepackage{hyperref}

\usepackage{orcidlink}

\usepackage[utf8]{inputenc}
\usepackage{amsmath}
\usepackage{amsfonts}
\usepackage{hyperref}
\usepackage{bbold}
\usepackage{parskip}
\usepackage{graphicx}
\usepackage{caption}
\usepackage{tikz}
\usepackage{amssymb}
\usepackage{todonotes}
\usepackage{wrapfig}

\begin{document}

\title{ScanTalk: 3D Talking Heads \\ from Unregistered Scans}
\author{Federico Nocentini$^*$\inst{1}
\and Thomas Besnier$^*$\inst{2}
\and  Claudio Ferrari\inst{4}
\and  Sylvain Arguillere \inst{5}
\and  Stefano Berretti\inst{1}
\and Mohamed Daoudi\inst{2,3}}
\authorrunning{F. Nocentini et al.}
%

\institute{
 Media Integration and Communication Center (MICC),\\ University of Florence, Italy\\ 
 \email{federico.nocentini@unifi.it, stefano.berretti@unifi.it}
\and Univ. Lille, CNRS, Centrale Lille, UMR 9189 CRIStAL, F-59000 Lille, France 
 \email{thomas.besnier@univ-lille.fr} \\ \and
IMT Nord Europe, Institut Mines-Télécom, Centre for Digital Systems
\email{mohamed.daoudi@imt-nord-europe.fr}\\
\and  Department of Architecture and Engineering University of Parma, Italy
\email{claudio.ferrari2@unipr.it}
\and Univ. Lille, CNRS, UMR 8524 Laboratoire Paul Painlevé, Lille, F-59000, France 
\email{sylvain.arguillere@univ-lille.fr}\\
 }



\maketitle
\def\thefootnote{*}\footnotetext{Equal contribution}

\begin{abstract}
Speech-driven 3D talking heads generation has emerged as a significant area of interest among researchers, presenting numerous challenges. Existing methods are constrained by animating faces with fixed topologies, wherein point-wise correspondence is established, and the number and order of points remains consistent across all identities the model can animate. In this work, we present \textbf{ScanTalk}, a novel framework capable of animating 3D faces in arbitrary topologies including scanned data. Our approach relies on the DiffusionNet architecture to overcome the fixed topology constraint, offering promising avenues for more flexible and realistic 3D animations. By leveraging the power of DiffusionNet, ScanTalk not only adapts to diverse facial structures but also maintains fidelity when dealing with scanned data, thereby enhancing the authenticity and versatility of generated 3D talking heads. Through comprehensive comparisons with state-of-the-art methods, we validate the efficacy of our approach, demonstrating its capacity to generate realistic talking heads comparable to existing techniques. While our primary objective is to develop a generic method free from topological constraints, all state-of-the-art methodologies are bound by such limitations.
Code for reproducing our results, and the pre-trained model are available at \href{https://github.com/miccunifi/ScanTalk}{https://github.com/miccunifi/ScanTalk}.
\keywords{3D Talking Heads \and 3D Scans Animation \and DiffusionNet}
\end{abstract}

\begin{figure}[h!]
    \centering
    \includegraphics[width=\linewidth]{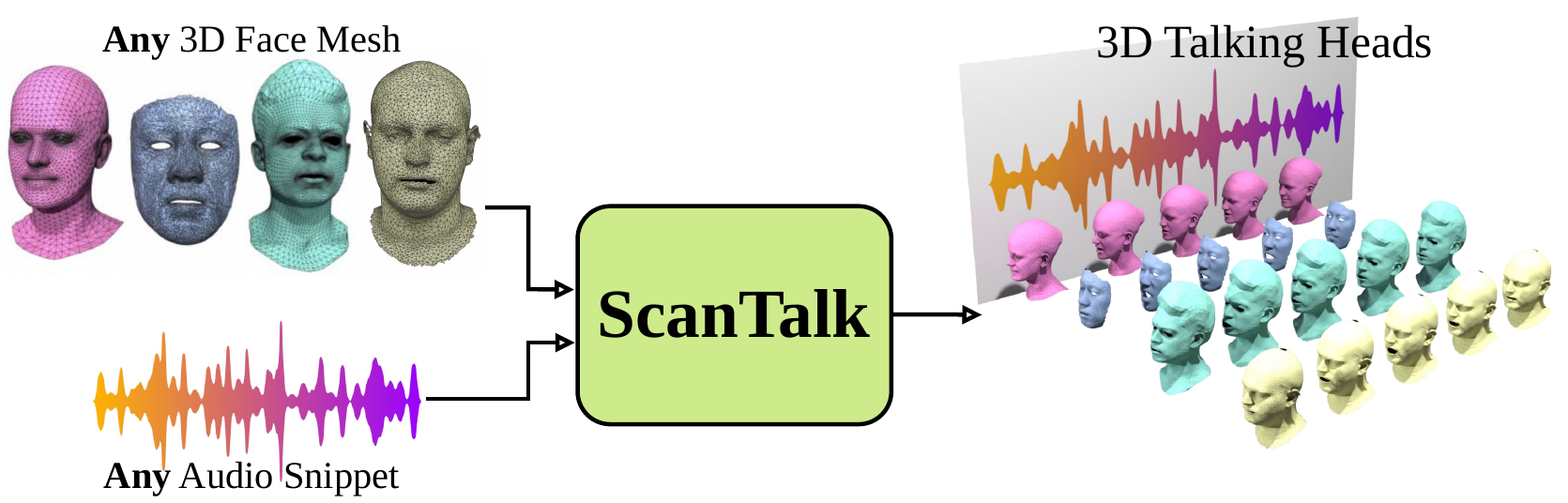}
    \caption{We present \textbf{ScanTalk}, a deep learning architecture to animate \textbf{any} 3D face mesh driven by a speech. ScanTalk is robust enough to learn on multiple unrelated datasets with a unique model, whilst allowing us to infer on unregistered face meshes.}
    \label{fig:intro_figure}
\end{figure}

\section{Introduction}
Human face animation is a complex task, widely explored in Computer Vision and Computer Graphics because of the broad range of applications drawn from it, spanning from virtual reality to video game graphics, and more. As 3D face models continue to improve, it becomes more and more relevant to incorporate multi-modal aspects such as speech with audio data. \textbf{Speech-driven facial animation} encounters challenges inherent to the fact that finding a cross-modality mapping from audio to geometric data is an ill-posed problem, dealing with the complex geometry of human faces and the limited availability of paired audio-visual 3D data. Another, less discussed aspect that touches a broader range of 3D graphics with meshes is the robustness to changes in the \textbf{topology} of the mesh, which refers to the arrangement of the vertices and how they are connected.

Maintaining fidelity and coherence across different topologies is crucial for ensuring that facial animations remain realistic and expressive, regardless of variations in the underlying mesh structure. This challenge becomes particularly pronounced in speech-driven facial animation, where the dynamics of speech-related lip movements, and related face changes necessitate a high degree of adaptability within the mesh topology.

Addressing these challenges requires innovative approaches able to navigate the complexities of facial geometry, while accommodating the nuances of speech-driven animation. In this regard, we present \textbf{ScanTalk}, a novel framework capable of animating faces in arbitrary topologies including scanned data. ScanTalk overcomes limitations associated with fixed topologies, offering promising avenues for more flexible and realistic 3D talking heads generation. Indeed, the aforementioned constraint limits the range of applications of current deep learning models for speech-driven motion synthesis. Because of this, deep models are required to train on large scale registered datasets such as FLAME~\cite{FLAME:SiggraphAsia2017} or Multiface~\cite{wuu2022multiface} for human face data. Building these datasets is a costly procedure, and the trained model is then only usable within the same registered setting. This typically means that any newly acquired data needs to be fitted to the corresponding topology before any animation can be predicted by a deep model. This extra preprocessing step usually prevents online applications, therefore canceling one of the key benefits of deep models, which is their inference speed.

Aiming to address these limitations, in this paper we present a flexible deep learning framework built to generate speech-driven animations of 3D face meshes. In particular, it gathers several key elements:
\begin{enumerate}
    \item A new robust approach to generate mesh deformation sequences based on DiffusionNet to compute intrinsic descriptors on 3D data;
    \item A comprehensive architecture for learning speech-driven animations. 
    Our approach works with meshes with different topologies, while showing competitive performance with respect to state-of-the-art models trained on an individual topology;
    \item We show the generalizability of our model to unseen mesh topologies with qualitative examples.
\end{enumerate}

\section{Related Work}
\label{sec:related_work}
In literature, there are several methodologies to acquire 3D face meshes, either by extracting the geometry from images or videos~\cite{yi2022_talkshow, mediapipe_2019} or by using a complex set of instruments in a controlled environment~\cite{bosphorus_2008, Texas3D_2010, BU3DFE_2006, COMA:ECCV18, BIWI_2010, wuu2022multiface}. The former offers advantages in terms of easier and more cost-effective data acquisition. However, these methods may sometimes fall short in capturing the complete 3D information from 2D data. With 3D scans, other challenges arise: they come unregistered and may present alterations such as holes and noise. Then, animating these objects directly becomes even more challenging. One can do it by registering the meshes onto a pre-defined topology~\cite{FLAME:SiggraphAsia2017, muralikrishnan_2023_BLISS, wuu2022multiface, BIWI_2010, Li_2020_Dynamic_facial_asset_and_rig_generation_from_a_single_scan} before animating the registrations with 3D morphable models~\cite{morphable_models_review} for example.

Several deep learning models~\cite{Liu_2019, Bahri_SMF_2021, Croquet_Diff_Reg_OT_2021, varifold_loss, shape_transformer} proposed ways to learn a latent representation of face scans using robust encoders such as PointNet~\cite{Charles_PointNet_2017, qi2017pointnet++} and Transformers but the resulting mesh is registered, which tends to smooth out some details from the scan geometry. However, this extra registration step may be handled efficiently with recent industrial applications such as MetaHuman from Epic Games.
More recently, and closer to our goal, DiffusionNet~\cite{sharp2021diffusion} was combined with neural Jacobian fields~\cite{NeuralJacobianField_2022} in Neural Face Rigging (NFR)~\cite{Qin_2023_NFR} to learn a per-triangle deformation field on faces, allowing to transfer an animation from a sequence of unregistered face meshes to another. 
We stand out from this work by proposing a model that uses audio data to animate a given unregistered face mesh. While DiffusionNet has demonstrated its capability to generalize across varying triangulations in a static setting, our model is the first to exhibit similar properties in a multi-modal 4D setting.

In recent years, numerous models and methodologies have emerged to address the challenge of synchronizing facial animations with speech audio. While significant progress has been made on 2D talking heads~\cite{EAMM_2022_2Dtalkinghead, Chen2020_2DtalkingheadGW, wang2021one_2Dtalkinghead, alghamdi2022_2Dtalking-head, Dipanjan_2020_2Dtalkinghead, Zhong_2023_CVPR_2Dtalkinghead, Wang2023Emotional_2Dtalkinghead}, only a handful of approaches can be seamlessly extended to 3D data.
Procedural techniques have been proposed to animate 3D faces~\cite{JALI_2016, dominance_Massaro_procedural_2001, Cosi_procedural_2002, Wang_2007_Rulebased_coarticulation_procedural, Xu_2013_diphone_coarticulation_procedural}, primarily relying on visemes (groups of phonemes) to drive the movements of facial muscles. However, many of these models necessitate re-targeting and struggle to generalize effectively without extensive processing steps.
More recent works leverage the increasing availability of data to develop statistical methods, including deep learning strategies~\cite{VOCA2019, richard2021meshtalk, Fan_Lin_Saito_Wang_Komura_faceformer_2022, facexhubert, Thambiraja_2023_ICCV_imitator, thambiraja2023_3diface, landmarks_3D_Nocentini_2023, xing2023codetalker, FaceDiffuser_Stan_MIG2023, nocentini2024emovocaspeechdrivenemotional3d}. 
Currently, state-of-the-art deep learning approaches are limited to a fixed mesh topology, requiring it to be identical to the one observed during the training phase. Among these models, VOCA~\cite{VOCA2019} pioneers the development of deep models trained on a large-scale dataset of registered meshes. It was outperformed when MeshTalk~\cite{richard2021meshtalk} introduced a larger registered dataset named Multiface~\cite{wuu2022multiface} along with a more expressive model. Moving forward, FaceFormer~\cite{Fan_Lin_Saito_Wang_Komura_faceformer_2022} utilized the transformer architecture, optimized for temporal data, and leveraged the power of a large-scale pre-trained audio encoder called Wav2vec2~\cite{wav2vec_2019}. This strategy saw further development with CodeTalker~\cite{xing2023codetalker} and SelfTalk~\cite{peng2023selftalk}. More recently, FaceXHubert~\cite{facexhubert} and FaceDiffuser~\cite{FaceDiffuser_Stan_MIG2023} utilized an improved audio encoder, Hubert~\cite{Hubert_audio_encoding}, demonstrating superiority to predict cross-modality mappings from audio to face motions. 
While continuously enhancing performances, these models are bound to a fixed mesh topology, hindering real-world applications and limiting the quantity of usable data for a single model, both for training and inference. 

In response to these challenges, we present a novel framework in the landscape of 3D facial animation learning, avoiding the limitations posed by the specific topology of the face mesh to be animated. The proposed model stands out for its capability to animate \textbf{any} face mesh, including real-world 3D scans. This novel approach holds the potential to redefine the standards in the field of 3D facial animation, being applicable across diverse datasets and scenarios.

\section{Proposed approach: ScanTalk}\label{sec:scantalk}
In this section, we introduce \textbf{Scantalk}, a framework for animating 3D face meshes reproducing a spoken sentence contained in an audio file, which does not require the meshes to adhere to any specific topology. 
With ScanTalk, we push the state-of-the-art in 3D talking heads a step forward by allowing any 3D face, even raw scans, to be animated given a speech. To train ScanTalk, the only requirement is that the training meshes share a common topology \textit{within each sequence}, but the topology may vary from one sequence to another. At inference time, any 3D face mesh in neutral state can be animated.

\begin{figure}[ht!]
    \centering
    \includegraphics[width=\linewidth]{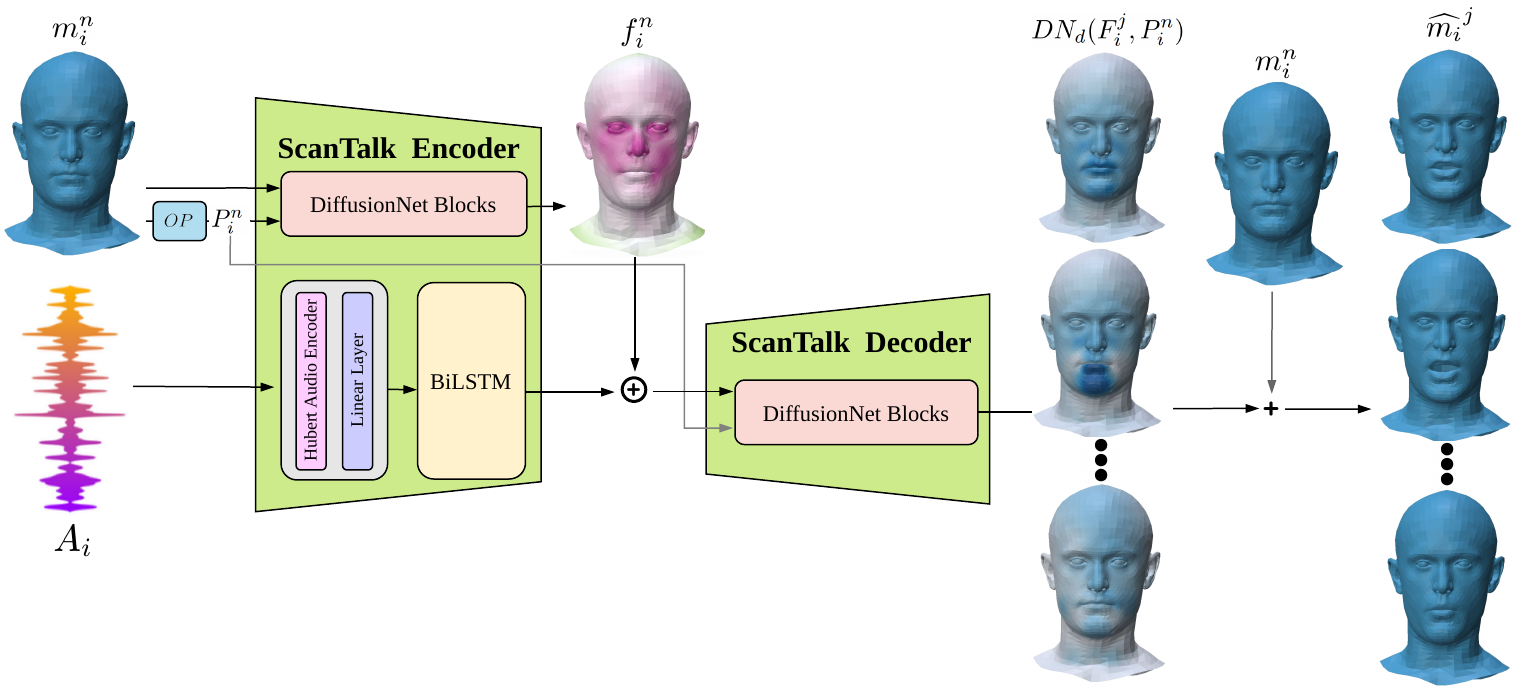}
    \caption{Architecture of \textbf{ScanTalk}. A novel Encoder-Decoder framework designed to dynamically animate any 3D face based on a spoken sentence from an audio file. The Encoder integrates the 3D neutral face $m_i^n$, per-vertex surface features $P_i^{n}$ (crucial for DiffusionNet and precomputed by the operators $OP$), and the audio file $A_i$, yielding a fusion of per-vertex and audio features. These combined descriptors, alongside $P_i^n$, are then passed to the Decoder, which mirrors a reversed DiffusionNet encoder structure. The Decoder predicts the deformation of the 3D neutral face, which is then combined with the original 3D neutral face $m_i^n$ to generate the animated sequence.}   
    \label{fig:lstm}
\end{figure}

ScanTalk is an Encoder-Decoder framework, which receives a neutral face mesh and an audio snippet as input, and outputs a sequence of per-vertex deformation fields, whose length depends on that of the audio. By summing each deformation field to the neutral input face, we ultimately obtain the animated sequence. The encoder is composed of two main modules: an audio encoder, which combines a pretrained encoder with a bi-directional LSTM that extracts audio features from the input speech, and a DiffusionNet encoder that computes surface descriptors from the neutral 3D face. These descriptors are then replicated and concatenated to the audio features and the resulting signal is fed to a DiffusionNet decoder that outputs the deformation to be applied to the neutral face. The framework is depicted in~\cref{fig:lstm}. Before providing the technical details in~\cref{sec:encoder} and~\cref{sec:decoder}, below we introduce a few general notations. 

Let $L = \left \{(M_i^{gt}, m_i^{n}, A_i)\right \}_{i=0}^{N-1}$ denotes the training set comprising $N$ samples, where $A_i$ is an audio containing a spoken sentence, $M_i^{gt} = (m_i^0, \dots, m_i^{T_i-1}) \in \mathbb{R}^{T_i \times V_i \times 3}$ represents a sequence of 3D faces (same topology) of length $T_i$ synchronized with the spoken sentence in $A_i$, and $m_i^{n}\in \mathbb{R}^{V_i \times 3}$ is a 3D neutral face. 
$V_i=\left |m_i^{n} \right |$ is the number of vertices in the $i$-th 3D face sequence that needs to be consistent across each mesh in the $i$-th sequence together with the vertex connectivity, overall determining the topology of the surface. 
Our objective is to establish a mapping function that correlates an audio input $A_i$, and a neutral 3D face $m_i^{n}$, to the ground-truth sequence $M_i^{gt}$, expressed as:
\begin{equation}
Scantalk(A_i, m_i^{n}) \approx M_i^{gt} .
\end{equation}

\subsection{ScanTalk Encoder}\label{sec:encoder}
ScanTalk requires an audio snippet $A_i$ and a 3D face in neutral state $m_i^n$ as inputs. To process the two inputs, ScanTalk comprises two distinct encoders to process the mesh and the audio, respectively, as detailed below.
\subsubsection{Face mesh encoder.} 
Several approaches are available for encoding face meshes or point clouds, yet traditional graph convolution-based models, like~\cite{spiralconv_Lim_2019, gong2019spiralnet++}, encounter limitations with varying graph structures, such as changes in mesh resolution. To address this, we employ DiffusionNet~\cite{sharp2021diffusion}, a discretization-agnostic encoder that demonstrated to be effective for encoding face meshes as seen in~\cite{Qin_2023_NFR}. DiffusionNet integrates multi-layer perceptrons (MLPs), learned diffusion, and spatial gradient features, offering a straightforward yet robust architecture for surface learning tasks. It bypasses the need for complex operations like explicit surface convolutions or pooling hierarchies.

Critical to DiffusionNet's functionality are precomputed features of the face surface, namely the \textit{Cotangent Laplacian}, \textit{Eigenbasis}, \textit{Mass Matrix}, and \textit{Spatial Gradient Matrix}. 
Together, these operators enhance the architecture's robustness, flexibility, and expressiveness, making DiffusionNet suitable for diverse surface learning tasks. Notably, this architecture accommodates 3D faces of any topology, allowing for variations in the number and order of defining points.\\
Let $P_i^n = OP(m_i^{n})$ represent the precomputed features obtained by applying the above surface closed-form operators $OP$ to the 3D neutral face $m_i^{n}$. 
The DiffusionNet Encoder $DN_e$, with latent size $h$, is designed to process a neutral 3D face mesh $m_i^{n}$, and requires the precomputed per-vertex features $P_i^n$ to extract a per-vertex descriptors $f_i^{n}$, that is:
\begin{equation}
f_i^{n} = DN_e(m_i^{n}, P_i^n) \in \mathbb{R}^{V_i \times h} .
\label{eq:per-vertex-descriptors}
\end{equation} 
The per-vertex descriptors $f_i^{n}$, is capable of capturing intricate details of each vertex within the neutral 3D face.

\subsubsection{Audio encoder.} Following~\cite{FaceDiffuser_Stan_MIG2023, facexhubert}, the speech encoder adopts the architecture of the state-of-the-art pretrained speech model, \textbf{HuBERT}~\cite{Hubert_audio_encoding} that is a self-supervised speech representation learner utilizing an offline clustering step to provide aligned target labels for a BERT-like prediction loss. Using this module, followed by a Linear Layer, we obtain a per-frame audio representation:
\begin{equation}
a_i = SpeechEncoder(A_i) \in \mathbb{R}^{T_i \times(h/2)} .
\end{equation}
To ensure coherence between the speech representation, following the methodology outlined in~\cite{landmarks_3D_Nocentini_2023}, we concatenate the \textit{SpeechEncoder} with a Multilayer Bidirectional-LSTM for temporal consistency, in a way such that the speech signal is projected into a \textbf{temporal latent representation}:
\begin{equation}
v_i = BiLSTM(a_i) \in \mathbb{R}^{T_i \times h} .
\end{equation}
\subsection{ScanTalk Decoder}\label{sec:decoder} 
We combine per-vertex descriptors $f_i^{n}$ extracted from the neutral face, with the latent vector $v_i$ extracted from the Bidirectional-LSTM
\begin{equation}
(F_i^j)_k = (f_i^{n})_k \oplus v_i^j \in \mathbb{R}^{h*2}, \hspace{1cm} \forall k = 0, \dots, V_i-1 ,\hspace{0,5cm} \forall j = 0, \dots, T_i-1 .
\end{equation}
With this concatenation, we obtain a combined latent $F_i^j \in \mathbb{R}^{V_i \times h*2}$ that embeds both audio-related and geometry-related latents. 
To decode this sequence of combined latents for deforming the neutral face $m_i^{n}$, we employ a DiffusionNet Decoder (which is essentially a reversed Encoder), denoted as $DN_d$. The decoder module receives $F_i^j$, and precomputed features $P_i^n$ derived from $m_i^{n}$. It predicts the deformation of $m_i^{n}$, denoted as $DN_d(F_i^j, P_i^n)$:
\begin{equation}
\widehat{m_i}^j = DN_d(F_i^j, P_i^n) + m_i^{n} \in \mathbb{R}^{V_i \times 3} .
\end{equation}
Here $\widehat{m_i}^j$ represents the $j$-th frame of the predicted sequence. The entire generated sequence is defined by $\widehat{M_i} \in \mathbb{R}^{T_i \times V_i \times 3}$.
We opt to utilize ScanTalk for predicting the deformation of the neutral face $m_i^n$ rather than predicting the actual face. This decision aligns with previous works~\cite{FaceDiffuser_Stan_MIG2023, Fan_Lin_Saito_Wang_Komura_faceformer_2022, peng2023selftalk, facexhubert, Thambiraja_2023_ICCV_imitator}, and offers advantages in terms of training efficiency and resulting animation. Predicting face deformation makes the learning process easier, by focusing solely on speech-related motion, as opposed to incorporating the problem of predicting the entire face reconstruction. Essentially, the decoder learns to predict a per-vertex displacement field from a time-dependent per-vertex descriptors field.

\subsection{ScanTalk Training} 
ScanTalk generates deformations of a neutral face; hence, a predicted sequence maintains a consistent topology across all frames.
This property arises from the definition of the DiffusionNet Decoder, which necessitates knowledge of precomputed features $P_i^n$, and the number of points in the 3D neutral face targeted for animation. 
For this reason, during a supervised training protocol, the ground-truth meshes within each sequence must adhere to a common topology. Despite the apparent specificity of this requirement, it proves to be non-prohibitive in practice.
ScanTalk animates a neutral face in response to an audio input, producing a sequence of 3D faces sharing the same topology of the neutral face that is animated, which can be any topology (even different from those seen during training).
This alignment between the training strategy and the desired inference outcome underscores the efficiency of ScanTalk, which is not restricted to the topology of the training faces. 

Notably, the model predicts per-vertex displacements of the neutral ground-truth mesh and the ground truth sequence is registered in a supervised setting. Consequently, the predicted sequence and the ground-truth sequence are aligned on the same topology but can vary from one sequence to another. Thus, during training, we minimize the average vertex-wise Mean Squared Error (MSE) over a sequence of length $T_i$ between the ground truth $M_i^{gt}$ and the model prediction $\widehat{M_i}$. which is defined as:
\begin{equation}
\mathcal{L}_{MSE} = \frac{1}{T_i-1}\sum_{j=0}^{T_i-1}\frac{1}{V_i-1}\sum_{k=0}^{V_i-1}\left \|(m_i^j)_k - (\widehat{m_i}^j)_k\right \|_2^2 .
\end{equation}
\section{Experiments}
In the following, we first introduce the datasets and the metrics used for evaluation, respectively in~\cref{sec:datasets} and~\cref{sec:metrics}. Then, we report quantitative and qualitative results in comparison with state-of-the-art methods in~\cref{sec:quantitative} and~\cref{sec:qualitative}.
In~\cref{sec:ablations} we present several ablation studies over the framework architecture. Finally, in~\cref{sec:user-study}, we report the results of an user study designed to compare our approach with state-of-the-art methods.
\subsection{Datasets}\label{sec:datasets}
For training and quantitative evaluations, we rely on three classical datasets in 3D speech-driven motion synthesis: VOCAset~\cite{VOCA2019}, BIWI~\cite{BIWI_2010} and Multiface~\cite{wuu2022multiface}.
\textbf{VOCAset} gathers mesh sequences of 12 actors performing 40 speeches, captured at 60fps. Each sequence lasts for around 3 to 5 seconds, and each mesh is registered to the FLAME~\cite{FLAME:SiggraphAsia2017} topology with 5,023 vertices and 9,976 faces.\\
\textbf{BIWI} comprises 14 subjects articulating 40 sentences each, sampled at 25fps. 
Each sentence lasts around 5 seconds, and the meshes are registered to a fixed topology. Due to GPU limitations, we used a downsampled version of this dataset, called BIWI$_6$, which has a fixed topology, with 3,895 vertices and 7,539 faces.\\
\textbf{Multiface} includes 13 identities executing up to 50 speeches of around 4 seconds each, sampled at 30fps. The original multiface meshes have a fixed topology with 5,471 vertices and 10,837 faces.

For training purposes, the meshes in both \textbf{Multiface} and \textbf{BIWI$_6$} have been scaled and aligned with the meshes in the \textbf{VOCAset} dataset. The data splits can be found in the supplementary material. Since ScanTalk is the first topology-independent 3D talking head generator, for the sake of a comprehensive comparison with the state-of-the-art, we trained 4 different models: one is trained using all the three datasets together (multi-dataset), while the other three are trained on each dataset separately (single-dataset). 

\subsection{Evaluation Metrics}\label{sec:metrics}
To evaluate the quality of the generated faces, we employ three standard metrics from previous works~\cite{richard2021meshtalk, xing2023codetalker, FaceDiffuser_Stan_MIG2023, peng2023selftalk}, namely:\\
\textbf{Lip vertex error} (LVE) $\times 10^{-5} (mm)$. Assesses the lip deviation in a sequence by comparing it to the ground truth. It is obtained by computing the maximum L2 error for all lip vertices in each frame, averaged across all frames.\\
\textbf{Mean vertex error} (MVE) $\times 10^{-3} (mm)$. Describes the quality of the overall face reconstruction by computing the maximal vertex L2 error for each frame and taking the mean across all frames.\\
\textbf{Upper-face dynamic deviation} (FDD) $\times 10^{-7} (mm)$. Compares the standard deviation of upper vertices between the generated sequence and ground-truth.\\

\subsection{Quantitative Results}\label{sec:quantitative}
\begin{figure}
    \centering
    \captionof{table}{\textbf{ScanTalk} performance, in both single-dataset (s-d) and multi-dataset (m-d) scenarios, in comparison with state-of-the-art methods. The heatmaps show the differences between the first frame and subsequent frames within sequences among the VOCAset, BIWI$_6$, and Multiface datasets. Notably, in the VOCAset, primarily the lips display movement, whereas in BIWI$_6$ and Multiface datasets, substantial head and upper face movements are observed. The color gradient on the face meshes corresponds to the average per-vertex $L_2$ norm of the differences, where blue hues indicate lower values, and red hues indicate higher values.}
  \begin{minipage}[t]{\linewidth}
    \hspace{2.6cm}
    \includegraphics[width=0.72\linewidth]{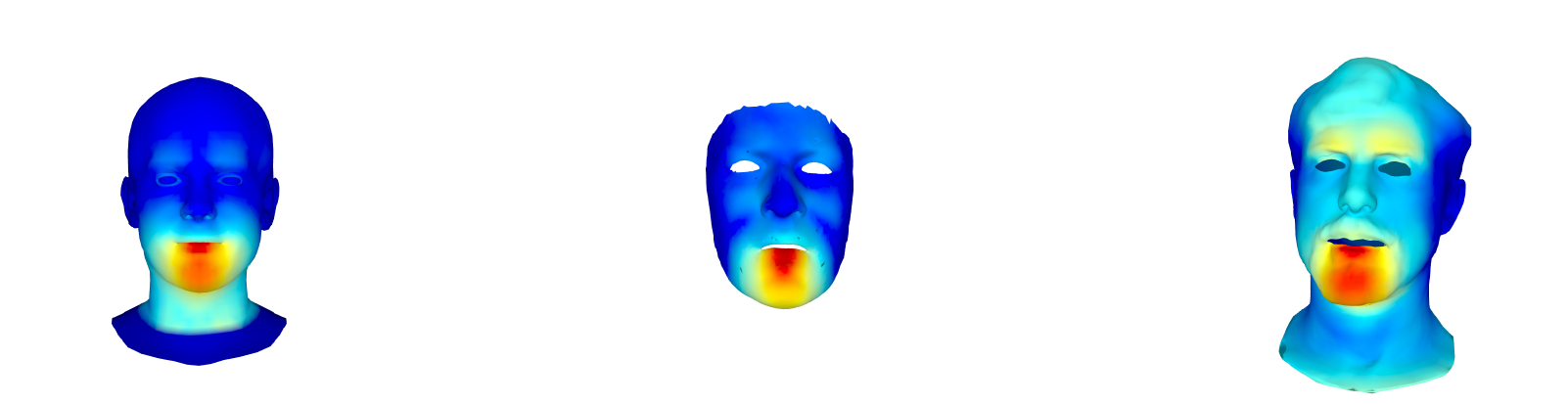}
  \end{minipage} 
  \begin{minipage}[t]{.94\linewidth}
  \centering
    \resizebox{\linewidth}{!}{
    \centering
    \begin{tabular}{@{}l@{\hspace{0.3cm}}ccccccccccc@{}} 
        \toprule
        & \multicolumn{3}{c}{\textbf{VOCAset}} & \phantom{abc} & \multicolumn{3}{c}{\textbf{BIWI}$_6$} &
        \phantom{abc} & \multicolumn{3}{c}{\textbf{Multiface}} 
        \\
        \cmidrule{2-4} \cmidrule{6-8} \cmidrule{10-12}
         & LVE $\downarrow$ & MVE $\downarrow$ & FDD $\downarrow$ && LVE $\downarrow$ & MVE $\downarrow$ & FDD $\downarrow$ && LVE $\downarrow$ & MVE $\downarrow$ & FDD $\downarrow$\\
        \midrule
        VOCA & 6.993 & 0.983 & 2.662 && 5.743 & 2.591 & 41.482 && 4.923 & 2.761 & 55.781 \\
        FaceFormer & 6.123 & 0.935 & \underline{2.163} && 4.085 & 2.163 & 37.091 && 2.451 & \textbf{1.453} & \textbf{20.239} 
        \\
        SelfTalk & 5.618 & 0.918 & 2.321 && \textbf{3.628} & \underline{2.062} & \underline{35.470} && \textbf{2.281} & 1.901 & 37.434 \\
        CodeTalker & \underline{3.549} & \underline{0.888} & 2.258 && 5.190 & 2.641 & \textbf{20.599} && 4.091 & 2.382 & 47.905 \\
        FaceDiffuser & 4.350 & 0.901 & 2.437 && \underline{4.022} & 2.128 & 39.604 && 3.555 & 2.388 & \underline{29.157} \\
        \midrule
        \textit{ScanTalk} m-d & 6.375 & 0.987 & \textbf{2.101} && 4.044 & \textbf{2.057} & 40.051 && \underline{2.435} & \underline{1.678} & 32.202 \\
        \midrule
        \textit{ScanTalk} s-d & \textbf{3.012} & \textbf{0.861} & 2.400 && 4.651 & 2.148 & 36.034 && 2.653 & 1.871 & 64.451 \\
        \bottomrule
        \end{tabular}
        }
  \end{minipage}
  \label{tab:comparison_table}
\end{figure}

Although ScanTalk is the first deep learning model to animate unregistered 3D face meshes, comparisons with the state-of-the-art is possible when inferring our model on registered data. 
In~\cref{tab:comparison_table}, we present the performance evaluation of ScanTalk training on both a single-dataset and multi-datasets, juxtaposed against several state-of-the-art methods trained solely on a single dataset. While the augmentation of training data enhances the effectiveness of our model, addressing disparities in data composition becomes important for animating unseen faces. Particularly, the varied geometries and head dynamics inherent in BIWI$_6$, VOCA, and Multiface meshes pose significant challenges, potentially limiting performance compared to a model trained exclusively on a single dataset.

This observation becomes evident when assessing Scantalk's performance on VOCAset, where a substantial improvement is observed when training on a single-dataset. Conversely, in the case of BIWI$_6$ and Multiface single-dataset training, a decline in results is noted compared to the multi-dataset version. This disparity is attributed to the presence of speech-related head and upper face movements within the ground truth talking head sequences of BIWI$_6$ and Multiface. Consequently, the multi-dataset training model demonstrates superior performance as it encounters a wider array of sequences featuring head and upper face movements. In contrast, VOCAset lacks such movements, with the 3D faces remaining static, while only the mouth moves, thereby enhancing performance for ScanTalk trained solely on VOCAset. This obviates the need to learn other movements of the head or upper face.

The distinctions between VOCAset and the other two datasets are further underscored by the FDD metric, indicating that 3D faces in VOCAset remain stationary, with minimal activity in the upper facial area and head. Across all tested models, generating sequences with lower FDD is notably more achievable on VOCAset compared to the other datasets, thus emphasizing the disparities between VOCAset and the others.

\begin{wrapfigure}{r}{0.35\textwidth}
     \centering
     \includegraphics[width=0.9\linewidth]{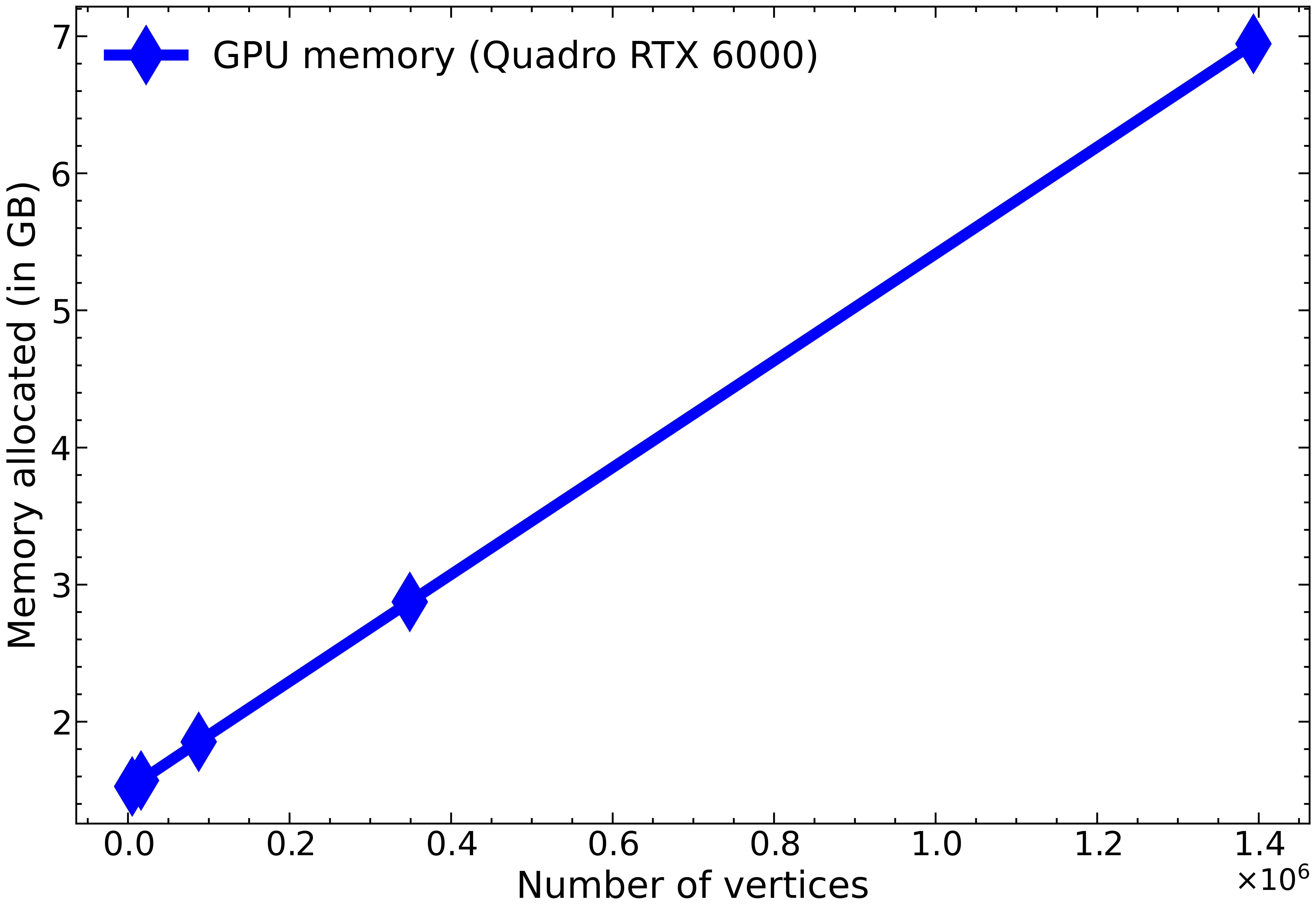}
     \caption{ScanTalk GPU memory usage with respect to the mesh resolution.}
     \label{fig:space_complexity}
\end{wrapfigure}
Nevertheless, the results presented in~\cref{tab:comparison_table} demonstrate that ScanTalk produces outcomes comparable to state-of-the-art methods, whether undergoing single-dataset or multi-dataset training. Moreover, ScanTalk emerges as the first model capable of training in multi-dataset settings, demonstrating the ability to animate a 3D face regardless of its topology.
An interesting characteristic of ScanTalk emerges when analyzing the GPU memory usage relative to the vertex count in the 3D facial model requiring animation. ~\cref{fig:space_complexity} illustrates a linear increase in GPU usage correlated with the number of vertices.

\subsection{Qualitative Results}\label{sec:qualitative}
In the domain of 3D speech-driven talking heads generation, qualitative evaluations hold as much significance as quantitative assessments. 
To evaluate the quantitative findings presented in~\cref{sec:quantitative}, we introduce a straightforward test aimed at discerning disparities in head and upper face movements across the three datasets: for each dataset, we compute the average $L_2$ norm of the differences between the initial frame and subsequent frames within each sequence. This enables the quantification of both head dynamics and upper facial movements across mesh sequences. In the heatmaps of~\cref{tab:comparison_table}, we present the per-frame average difference for each dataset. Our analysis reveals that in the VOCAset, only the lower part of the face exhibits discernible movements, whereas in BIWI$_6$ and Multiface, the entire head or face moves.
\begin{figure}[h!]
\centering
\includegraphics[width=0.9\linewidth]{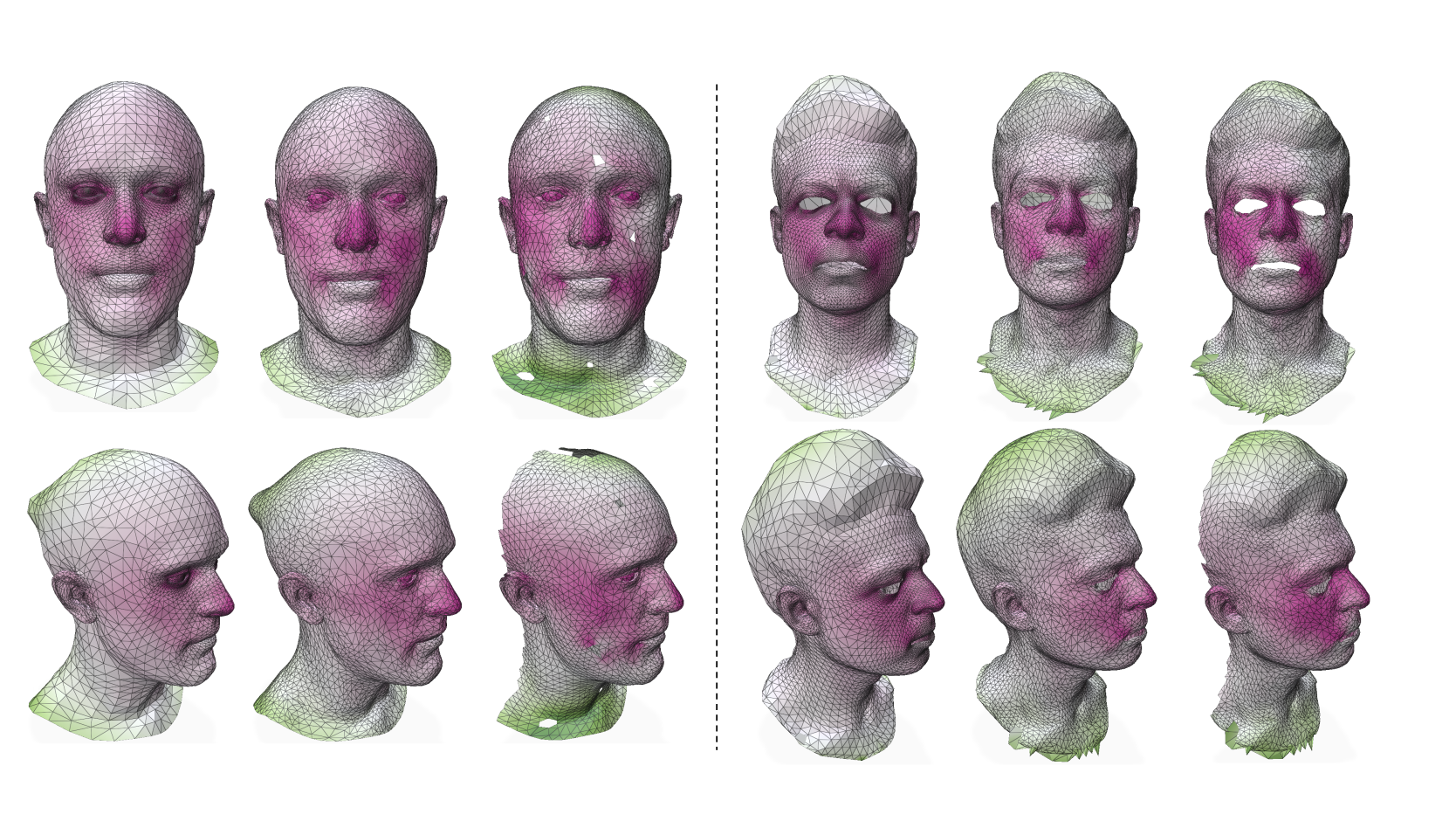}
\caption{Relative norm of the per-vertex descriptors $f_i^n$ in~\cref{eq:per-vertex-descriptors} extracted by $DN_e$ displayed as a heatmap on a mesh from VOCAset (left), and a mesh from Multiface (right). For each mesh, we show the norm on the original topology, on a remeshed version, and on a further degraded mesh obtained by removing the back of the head and creating random holes. Here, pinker hues indicate lower values, and greener hues indicate higher values.}
\label{fig:norms}
\end{figure}
\begin{figure}[ht!]
\centering
\includegraphics[width=\linewidth]{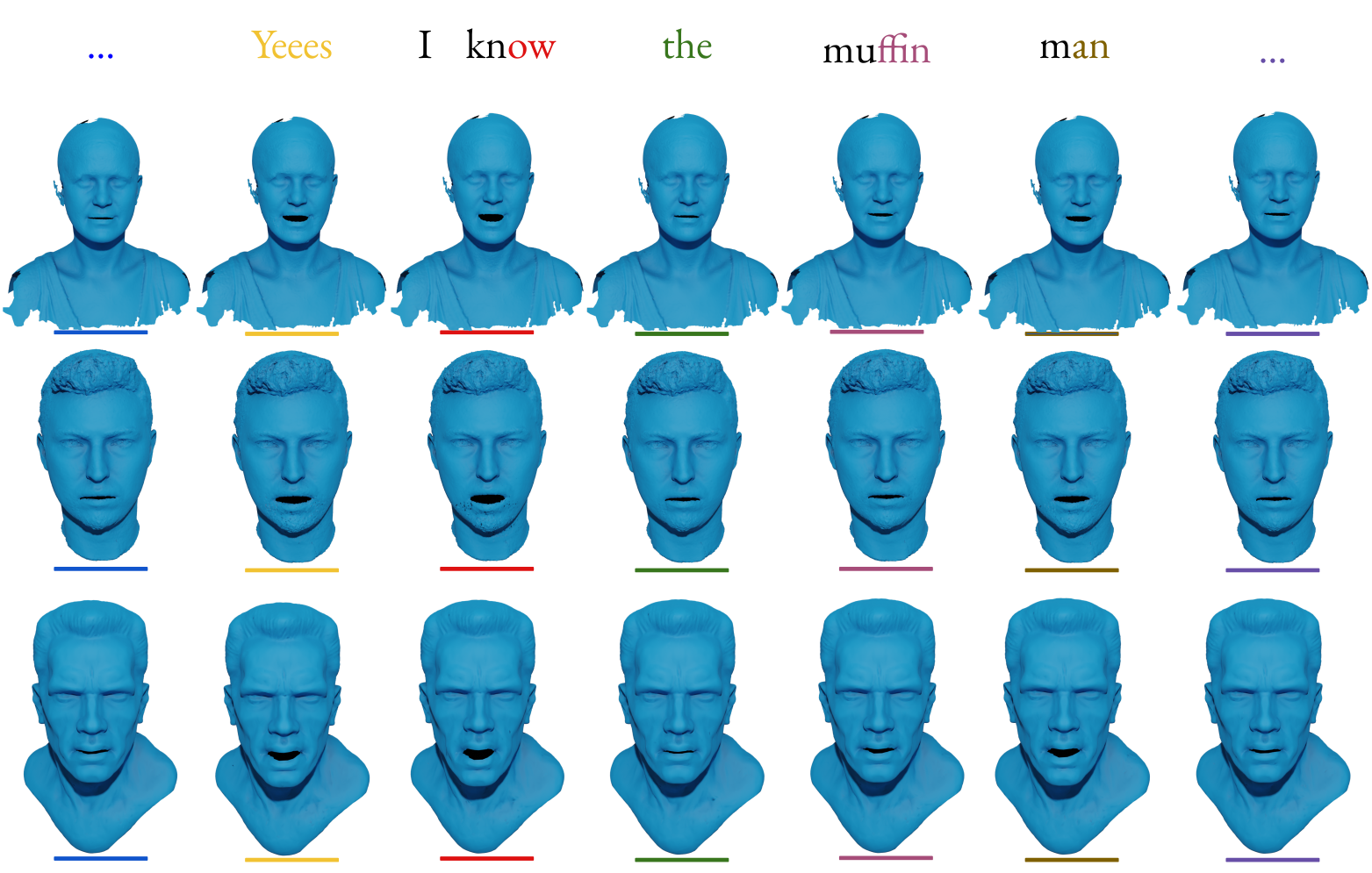}
\caption{ScanTalk inference on 3D faces. The meshes have been rigidly aligned with the training data. The first row is an animation of a raw 3D scan (an hole for the mouth has been created), while the others two are animation of meshes in an arbitrary topology.}
\label{fig:scans}
\end{figure}

In~\cref{fig:norms}, we present a heatmap visualization representing the norm of geometric descriptors $f_i^n$ computed by the encoder $DN_e$ in~\cref{eq:per-vertex-descriptors} for various alterations of the same identity mesh. Our findings indicate that the ScanTalk encoder effectively extracts descriptors from input faces regardless of facial topology. Notably, despite differences in facial topology among the same identity, the extracted descriptors exhibit remarkable similarity.

In~\cref{fig:scans}, 3D faces in different topology, including a scan, animated using ScanTalk are presented. Our method demonstrates good generalization capabilities, successfully animating diverse 3D faces. Notably, enhanced animation quality is observed when a mouth aperture is present; however, our model performs well even in its absence. Nonetheless, without a mouth aperture, the model struggles to generate the corresponding mouth opening, despite accurate lip movement synchronization. 

\subsection{Ablations Studies}\label{sec:ablations}
To investigate the impacts of various components within our architecture, we conducted extensive experiments across various configurations, encompassing both single-dataset and multi-dataset training paradigms.
The results reported in~\cref{tab:singledataset-ablation} and~\cref{tab:multidataset_ablation} are computed by modifying ScanTalk's modules and training losses as described in~\cref{sec:scantalk}.
\subsubsection{Audio encoder.} 
In learning based speech-driven animation, some previous work used either Wav2vec2~\cite{wav2vec_2019} or Hubert~\cite{Hubert_audio_encoding}, the latter being substantially better for this task according to~\cite{FaceDiffuser_Stan_MIG2023, facexhubert}. 
For our purpose, we also tested WavLM~\cite{Chen2021WavLMLS} but obtained poorer efficiency. The results on both the single-dataset and multi-dataset training are displayed in~\cref{tab:compare_audio_encoders} and~\cref{tab:multidataset_ablation} top. 
We can see that, in both cases, the usage of the Hubert Encoder~\cite{facexhubert} for audio features extraction leads to better results.
\subsubsection{Temporal consistency.}
In our investigation, as proven by~\cite{Thambiraja_2023_ICCV_imitator, landmarks_3D_Nocentini_2023, facexhubert}, we found that supplementing pre-trained audio encoders with additional temporal consistency mechanisms such as Bidirectional-LSTM, Bidirectional-GRU, or an autoregressive Transformer Decoder (TD) significantly enhances model performance, as illustrated in~\cref{tab:temporal_consistency_comparison} and in~\cref{tab:multidataset_ablation} middle. 
The BiLSTM model architecture is detailed in~\cref{sec:scantalk}, while the BiGRU model mirrors the BiLSTM architecture but substitutes BiLSTM with BiGRU. 
Details about the Transformer Decoder can be found in the supplementary material.
From the findings presented in~\cref{tab:temporal_consistency_comparison} and in~\cref{tab:multidataset_ablation} middle, we observe that employing a multilayer Bidirectional-LSTM in the audio stream processing of ScanTalk yields the most favorable performance across both single-dataset training and multi-dataset training scenarios.

\subsubsection{Loss function.}
Previous research has extensively explored optimal objective functions for enhancing and refining the learning process. 
Inspired by~\cite{Thambiraja_2023_ICCV_imitator, landmarks_3D_Nocentini_2023, peng2023selftalk}, we experimented with Mean Square Error (\textbf{$L_{MSE}$}), Masked (\textbf{$L_{mask}$}), and Velocity (\textbf{$L_{vel}$}) Loss. 
The former is a simple $L_2$ loss, the second employs Mean Square Error focused on lip vertices, while the latter aims to minimize differences between consecutive frames.
While these loss functions have demonstrated efficacy in single-dataset training, as evidenced by~\cite{Thambiraja_2023_ICCV_imitator, landmarks_3D_Nocentini_2023, peng2023selftalk}, our findings, detailed in~\cref{tab:multidataset_ablation} bottom, indicate that employing a straightforward $L_{2}$, on multi-dataset training, enhances the model's capacity to generate realistic talking heads. We attribute this to significant geometric variations observed across different datasets.

\begin{table}[!ht]
\caption{ScanTalk (ST) single-dataset ablation studies. Results obtained with a model trained just on VOCAset.} 
\resizebox{\linewidth}{!}{
\begin{subtable}[t]{0.6\linewidth}
\centering
\caption{\normalsize Audio Encoder Ablation.}
\label{tab:compare_audio_encoders}
\begin{tabular}{l@{\hspace{0.5cm}}ccc}
\toprule
         & LVE $\downarrow$ & MVE $\downarrow$ & FDD $\downarrow$ \\
\midrule
ST w WavLM& 3.674 & 0.937 & 2.413\\
ST w Wav2Vec2& \underline{3.309} & \textbf{0.860} & \textbf{2.244}\\ 
\textit{ScanTalk}& \textbf{3.012} & \underline{0.861} &     \underline{2.400} \\
\bottomrule
\end{tabular}
\end{subtable}
\begin{subtable}[t]{0.6\linewidth}
\centering
\caption{\normalsize Temporal consistency ablation.}
\label{tab:temporal_consistency_comparison}
\begin{tabular}{l@{\hspace{0.5cm}}ccc}
\toprule
         & LVE $\downarrow$ & MVE $\downarrow$ & FDD $\downarrow$ \\ 
\midrule
ST w/o    & 3.361 & 0.870 & \underline{2.365}\\
ST w TD& 3.291 & \underline{0.859} & 2.406\\
ST w BiGRU& \underline{3.036} & \textbf{0.835} & \textbf{2.358}\\
\textit{ScanTalk}& \textbf{3.012} & 0.861 & 2.400 \\
\bottomrule
\end{tabular}
\end{subtable}
}
\label{tab:singledataset-ablation}
\end{table}
\begin{table*}[h!]
\centering
\caption{ScanTalk (ST) multi-dataset ablation studies. \textbf{Top}: ST with different audio encoders. \textbf{Middle}: ST using different audio stream processing. \textbf{Bottom}: ST with different Loss function. The last row is the proposed ScanTalk}
\label{tab:multidataset_ablation}
\small
\resizebox{0.94\linewidth}{!}{
\begin{tabular}{@{}l@{\hspace{0.3cm}}ccccccccccc@{}} 
\toprule
& \multicolumn{3}{c}{\textbf{VOCAset}} & \phantom{abc} & \multicolumn{3}{c}{\textbf{BIWI}$_6$} &
\phantom{abc} & \multicolumn{3}{c}{\textbf{Multiface}} 
\\
\cmidrule{2-4} \cmidrule{6-8} \cmidrule{10-12}
& LVE $\downarrow$ & MVE $\downarrow$ &  FDD $\downarrow$ && LVE $\downarrow$ & MVE $\downarrow$ & FDD $\downarrow$ && LVE $\downarrow$ & MVE $\downarrow$ & FDD $\downarrow$\\
\midrule
ST w Wav2Vec2& 7.127 & 1.241 & 2.349 && 4.439 & 2.276 & 41.374 && 3.198 & 2.329 & 33.671 \\
ST w WavLM& 6.773 & 1.004 & 2.220 && \underline{4.333} & 2.162 & \textbf{31.922} && 2.907 & 2.004 & \textbf{8.500} \\
\midrule
ST w TD& \underline{6.512} & 0.995 & 1.858 && 4.827 & 2.168 & 36.789 && \underline{2.136} & \underline{1.725} & 35.044 \\
ST w  BiGRU& 6.584 & \underline{0.994} & 2.054 && 4.421 & \underline{2.103} & 41.456 && 2.534 & 1.987 & 34.098 \\
\midrule
ST$+L_{mask}+L_{vel}$& 7.031 & 1.092 & \underline{1.473} && 4.755 & 2.227 & 39.458 && 2.547 & 1.898 & \underline{15.973} \\
ST$+L_{mask}$& 7.451 & 1.084 & \textbf{0.859} && 4.748 & 2.199 & 36.939 && 2.888 & 1.891 & 32.501 \\
ST$+L_{vel}$& 6.740 & 0.998 & 1.899 && 4.509 & 2.180 & \underline{33.614} && \textbf{2.103} & 1.775 & 31.599 \\
\midrule
\midrule
\textit{ScanTalk}& \textbf{6.375} & \textbf{0.987} & 2.101 && \textbf{4.044} & \textbf{2.057} & 40.051 && 2.435 & \textbf{1.678} & 32.202 \\
\bottomrule
\end{tabular}
}
\end{table*}
\subsection{User study}\label{sec:user-study}
To further evaluate our solution, we performed a study where human feedback is involved. This evaluation is conducted with two user-based studies involving 25 participants; \emph{(i)} For the first study, in alignment with prior research~\cite{peng2023selftalk, xing2023codetalker, Fan_Lin_Saito_Wang_Komura_faceformer_2022, facexhubert, FaceDiffuser_Stan_MIG2023, richard2021meshtalk}, we designed an A/B test to compare ScanTalk with other state-of-the-art models within a registered setting on both lip-syncing and naturalness criteria (Test~1); \emph{(ii)} In the second test, we assessed the credibility of scan animations by asking participants to rate the animation quality of ten scans sourced from the COMA dataset~\cite{COMA:ECCV18}, using a scale ranging from 1 to 10 (Test~2).

\begin{figure}[!ht]
  \begin{minipage}[t]{.42\linewidth}
  \vspace{-3.3cm}
    \resizebox{\linewidth}{!}{
        \begin{tabular}{lc@{\hspace{0.2cm}}c}
        \toprule
        \textbf{Comparison} & \textbf{Naturalness (\%)} & \textbf{Lip-sync (\%)} \\ 
        \midrule
       ST vs. FaceFormer   & 82.67 & 80.67 \\
       ST vs. CodeTalker   & 57.33 & 56.00 \\
       ST vs. FaceDiffuser & 90.67 & 94.00 \\
       ST vs. SelfTalk     & 44.00 & 43.33 \\ 
       ST vs. GT           & 44.00 & 42.00 \\
        \bottomrule
        \end{tabular}
        }
  \end{minipage}
  \begin{minipage}[t]{.6\linewidth}
    \includegraphics[width=\linewidth]{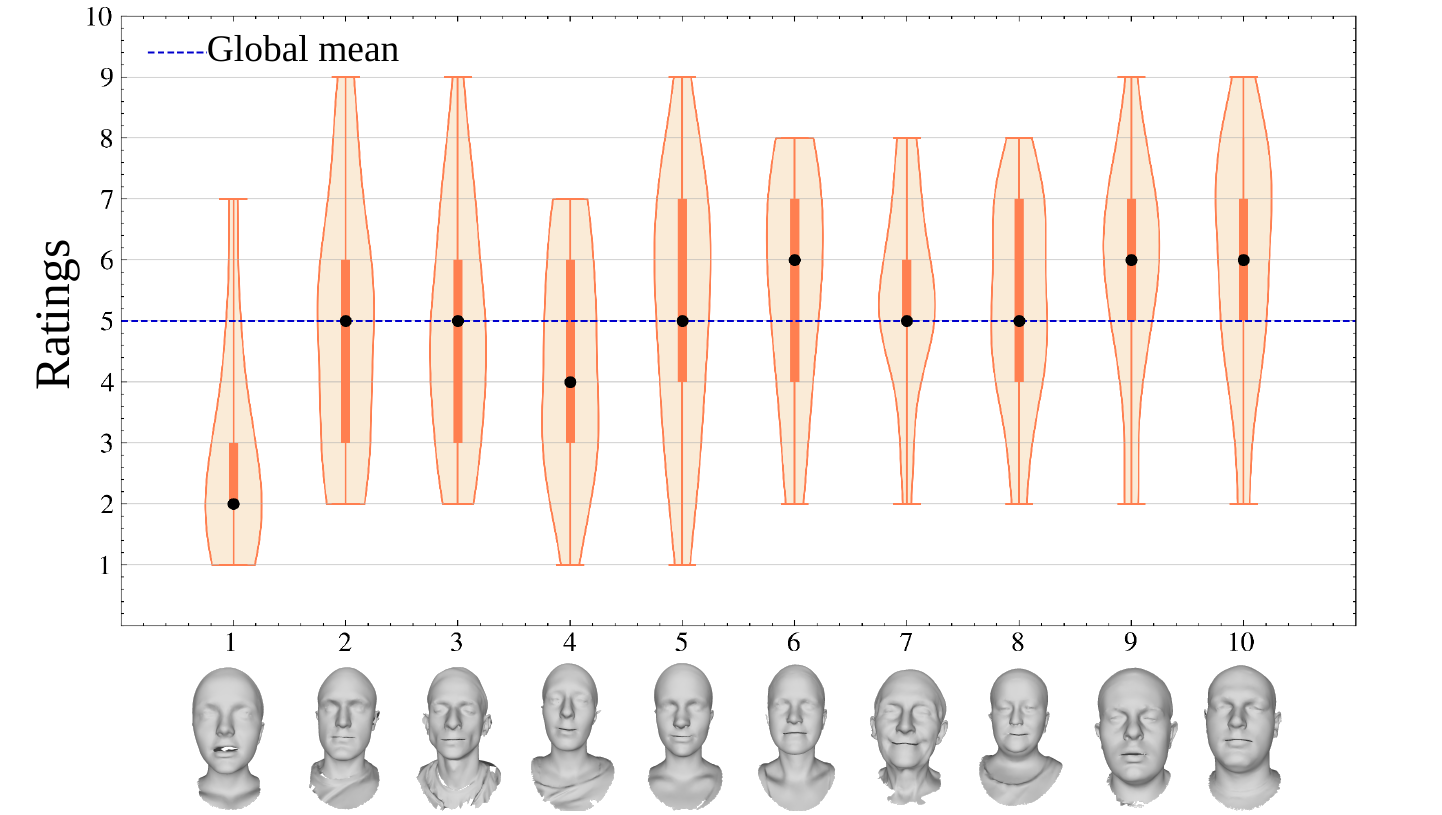}
  \end{minipage} 
  \captionof{figure}{The table on the left shows results of Test~1, ScanTalk (ST) vs. other methods and the ground truth (GT) on samples from the test set of VOCAset. The values denote the percentages of users who favored ScanTalk animations over others. The chart on the right shows results of Test~2 reported as a violin plot for the animation of scans from COMA~\cite{COMA:ECCV18}. The subject median rating is displayed as a black dot on each violin.}
  \label{fig:user_study}
\end{figure}

The outcomes of Test~1 are depicted in the table on the left of~\cref{fig:user_study}. Notably, sequences produced using ScanTalk demonstrate levels of both naturalness and lip-syncing that are comparable to state-of-the-art methods. Specifically, our approach is preferred when compared with FaceFormer, CodeTalker, and FaceDiffuser on VOCAset. However, SelfTalk and the ground truth are preferred over ScanTalk. Nevertheless, the percentage of users favoring ScanTalk remains non-negligible, underscoring the efficacy of our approach in generating 3D talking heads with good realism and lip-syncing fidelity.

Results of Test~2 are reported on the right of~\cref{fig:user_study}. These ratings are closely linked to both the scan quality and the fidelity of the mouth representation. Despite the scan quality not being optimal, as depicted in~\cref{fig:user_study}, our results indicate that we can achieve a good level of realism in the animated scans.
\section{Conclusions}
This paper introduces ScanTalk, a novel framework for speech-driven 3D facial animation. Unlike existing methods, ScanTalk possesses the unique capability to animate any 3D face, regardless of its topology, even if it differs from the ones on which it was trained. ScanTalk extends the applicability of deep speech-driven 3D facial animations by addressing the challenges of topology robustness. Additionally, our model demonstrates comparable quantitative and qualitative results with other state-of-the-art methods across three distinct datasets.\\
\textbf{Acknowledgments}
This work is supported by the ANR project Human4D ANR-19-CE23-0020 and by the    \href{https://geogen3dhuman.univ-lille.fr}{IRP CNRS project GeoGen3DHuman}. This work was also partially supported by ``Partenariato FAIR (Future Artificial Intelligence Research) - PE00000013, CUP J33C22002830006" funded by NextGenerationEU through the italian MUR within NRRP, project DL-MIG. This work was also partially funded by the ministerial decree n.352 of the 9th April 2022 NextGenerationEU through the italian MUR within NRRP. This work was also partially supported by Fédération de Recherche Mathématique des Hauts-de-France (FMHF, FR2037 du CNRS).

 
 \section{Supplementary Material}

In this supplementary material, we provide additional details and results that did not fit into the main paper. 

\subsection{Ethical Comments}
We recognize the ethical considerations surrounding the creation of 3D facial animations. Generating synthetic narratives with 3D faces poses inherent risks, potentially resulting in both intentional and unintentional consequences for individuals and society as a whole. We emphasize the importance of adopting a human-centered approach in the development and implementation of such technology.

\subsection{Transformer Decoder}\label{sec:td}
\begin{figure}[ht!]
    \centering
    \includegraphics[width=\linewidth]{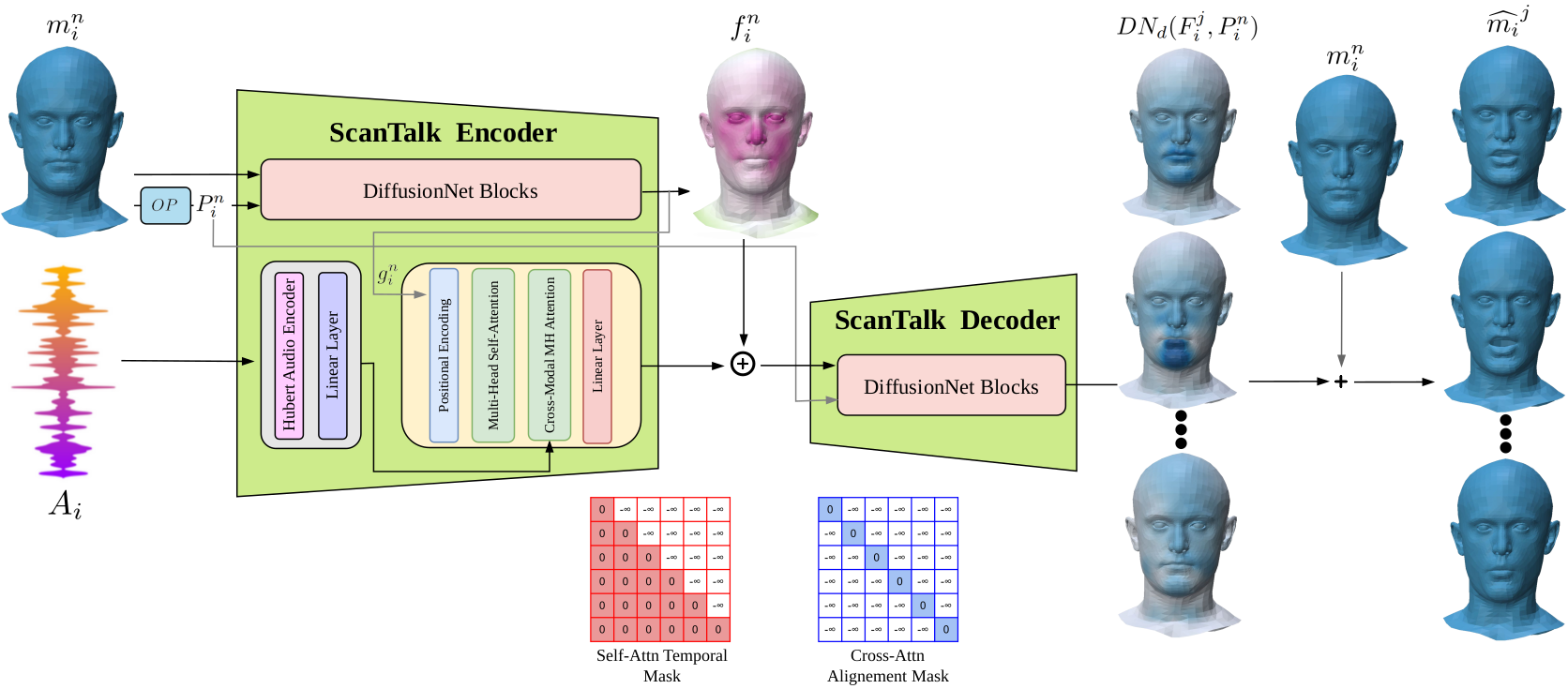}
    \caption{Architecture of \textbf{ScanTalk} Transformer.}   
    \label{fig:transformer}
\end{figure}

Inspired by prior works ~\cite{Fan_Lin_Saito_Wang_Komura_faceformer_2022, Thambiraja_2023_ICCV_imitator}, ScanTalk with Transformer Decoder  follows a distinct approach, described in~\cref{fig:transformer}. It employs a \textit{SpeechEncoder} module preceding an autoregressive Transformer Decoder, which necessitates an initial token. Unlike traditional methods, our approach initializes the generation process with the global representation of $m_i^{neutral}$, the neutral face for animation, denoted as $g_i^n$, serving as the starting token. The per-vertex features are aggregated through averaging, yielding:

\begin{equation}
g_i^n = \dfrac{1}{V_i}\sum\limits_{k=1}^{V_i} (f_i^n)_k \in \mathbb{R}^{h}.
\end{equation}

This global feature vector, $g_i^n$, encapsulates fundamental attributes of the neutral face, providing valuable insights into its overall structure and characteristics. While Faceformer ~\cite{Fan_Lin_Saito_Wang_Komura_faceformer_2022} commences generation with an embedding of a one-hot label representing the speaker, and Imitator ~\cite{Thambiraja_2023_ICCV_imitator} begins from a zero token, our methodology offers a novel perspective on initializing the generation process.

The Transformer Decoder comprises a concatenation of components: a \textit{Positional Encoding Layer} encoding token positions in the sequence, a \textit{Masked Self-Attention Layer} incorporating information from preceding tokens, and a \textit{Masked Cross-Attention Layer} combining token information with corresponding details from the \textit{SpeechEncoder}. The autoregressive token generation process is defined as:

\begin{equation}
    v_i^j = TD(v_i^{1:j-1}, a_i^j) \in \mathbb{R}^{h} \hspace{1cm} \forall j = 1, \dots, T_i \hspace{0.4cm} with \hspace{0.4cm} v_i^0 = g_i^n.
\end{equation}

\subsection{Implementation details}\label{sec:imp_details}
Our ScanTalk model, as described in Section 3 of the main paper, is constructed as follows: the DiffusionNet Encoder comprises 4 DiffusionNet blocks, each with a hidden size ($h$) of 32. The Bi-LSTM consists of 3 layers with a hidden size of 32. The DiffusionNet Decoder accepts as input the concatenation of features of dimension 64 and outputs the per-vertex deformation of the neutral face. \\
The DiffusionNet Decoder is composed of 4 DiffusionNet blocks concatenated together.
All the ScanTalk versions presented in the main paper are trained for 200 epochs over each dataset using the Adam optimizer~\cite{adam}, with a learning rate of $10^{-4}$.

\subsection{Datasets}\label{sec:dataset_split}
\begin{figure}[!ht]
\centering
  \begin{subfigure}[b]{.4\linewidth}
  \includegraphics[width=\linewidth]{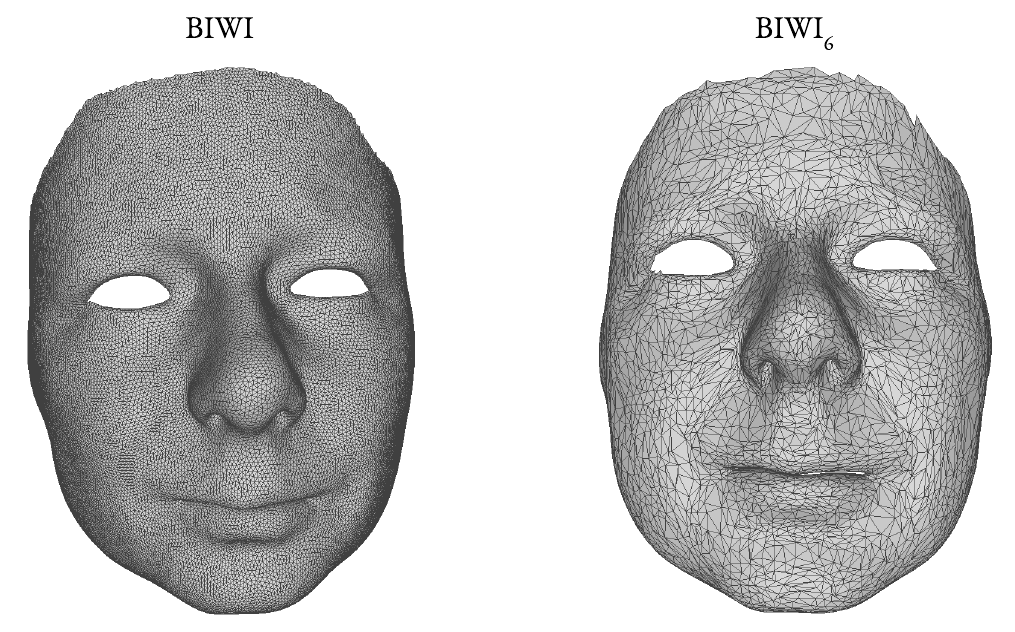}
  \label{fig:BIWI_VS_BIWI_6}
  \end{subfigure}
  \vline
  \begin{subfigure}[b]{.4\linewidth}
    \includegraphics[width=\linewidth]{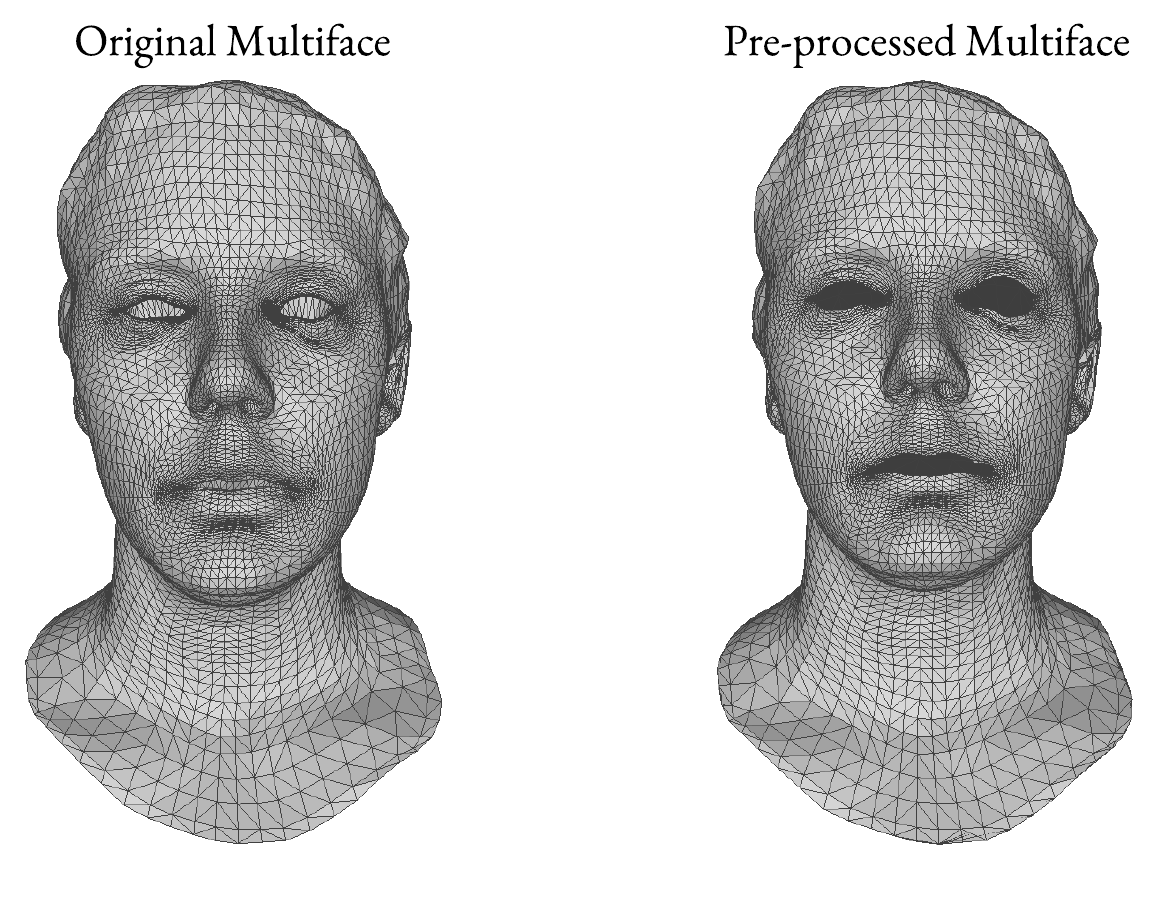}\label{fig:original_multiface_VS_processed_multiface}
  \end{subfigure} 
  \captionof{figure}{(Left) Side by side comparison of an original mesh from BIWI and the same mesh in BIWI$_6$. (Right) Side by side comparison of an original mesh from Multiface and the same mesh after preprocessing.}
  \label{fig:comparison}
\end{figure}

We summarize the characteristics of the datasets in ~\cref{tab:datasets_specs}. Our preprocessing of the BIWI dataset is depicted in ~\cref{fig:comparison} Left, while the manipulation applied to the Multiface dataset is illustrated in~\cref{fig:comparison} Right. Specifically, the BIWI dataset underwent downsampling and rigid alignment with the VOCAset, whereas the Multiface dataset was rigidly aligned with the VOCAset, with additional modifications involving the creation of three apertures corresponding to the eyes and mouth.

\begin{table}[h!]
\centering
\caption{Train / test / val splits for each dataset.}
\label{tab:datasets_specs}
\begin{tabular}{l@{\hspace{0.5cm}}l@{\hspace{0.3cm}}l@{\hspace{0.3cm}}l@{\hspace{0.3cm}}l}
\toprule
         & \textbf{VOCAset} & \textbf{BIWI}$_6$ & \textbf{Multiface}  \\ 
\midrule
Type                 & Head + neck & Narrow face & Head + neck \\
\# vertices          & 5,023 & 3,895 & 5,471  \\
\# faces             & 9,976 & 7,539 & 10,837 \\
Training samples     & 320 & 400 & 410  \\ 
Val samples          & 80 & 80 & 100 \\
Test samples         & 80 & 80 & 100 \\
\end{tabular}
\end{table}

\subsection{Mesh encoding}\label{sec:mesh_enc}
Several encoding strategies for geometry are feasible; however, with our experimentation we saw that encoding vertex positions provides the optimal and most intuitive approach. When omitting mesh encoding and directly feeding the BiLSTM output to the decoder, the mesh signal remains constant across frames, leading to a static facial expression as the decoder lacks spatial awareness of the mouth's location. Incorporating normals alongside positions fails to enhance results, as precomputed operators already furnish adequate orientation information. Additionally, adopting the Heat Kernel Signature (HKS), as suggested in the DiffusionNet framework, does not yield improvements in results. In~\cref{fig:norms}, we present the per-vertex norm of features derived from the DiffusionNet Encoder for both training and testing meshes.

\begin{figure}[!h]
    \centering
    \includegraphics[width=0.8\linewidth]{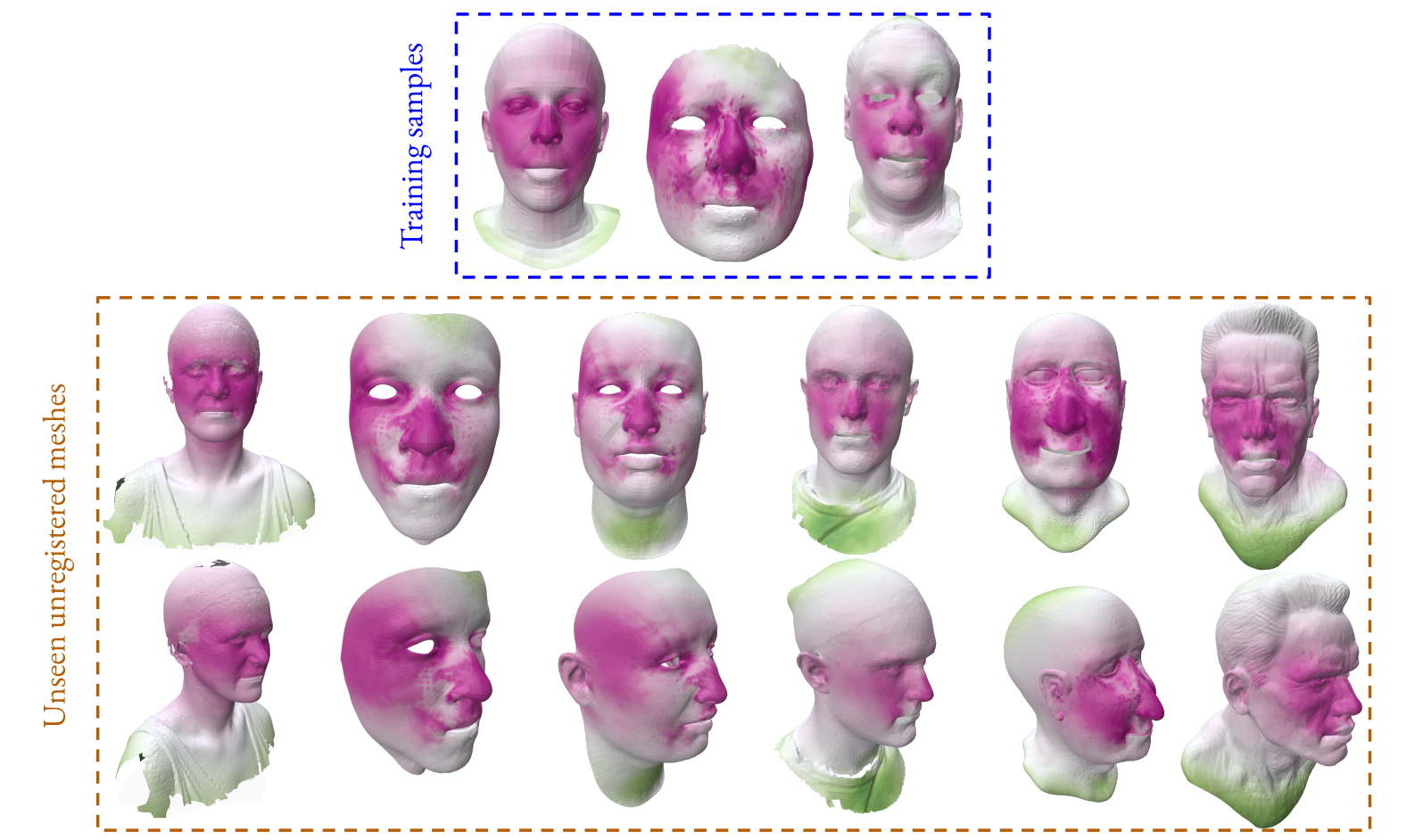}
    \caption{Relative norm of the per-vertex descriptors extracted by the encoder displayed as a heatmap where pinker hues indicates lower values and greener hues indicates higher values.}   
    \label{fig:norms}
\end{figure}

\subsection{Additional Qualitative results}\label{sec:add_qual}
\begin{figure}[!h]
    \centering
    \includegraphics[width=0.95\linewidth]{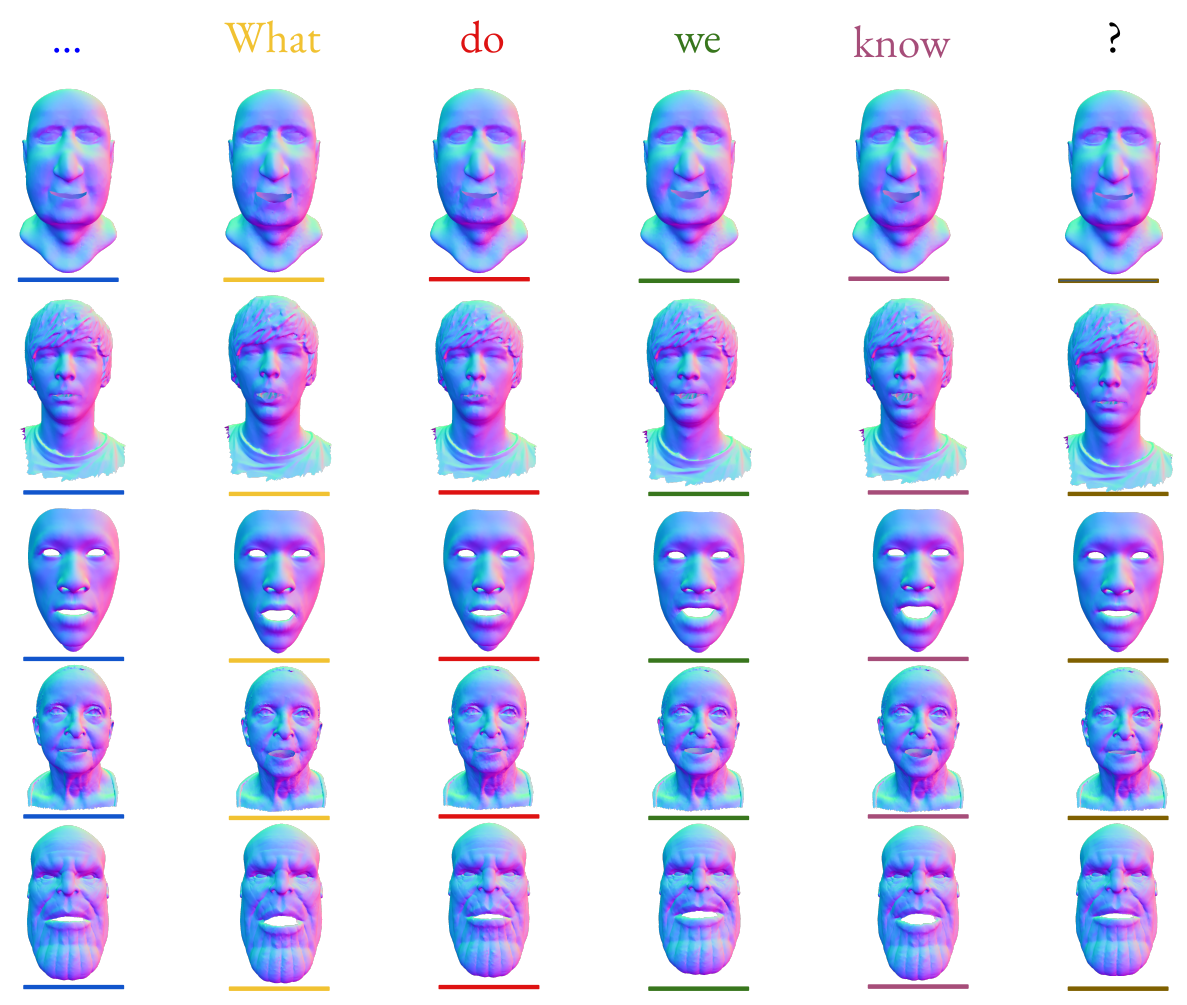}
    \caption{Additional experiments with different unseen meshes.}   
    \label{fig:more_exp}
\end{figure}

In~\cref{fig:more_exp}, we present qualitative examples of animation using ScanTalk applied to 3D faces with arbitrary topology. Our preprocessing steps included rigid alignment with training meshes and the creation of an aperture for the mouth. 
From~\cref{fig:more_exp} it is evident that ScanTalk exhibits a remarkable capacity for generalization, enabling animation of any 3D face once aligned with the training set and provided with a mouth aperture. Notably, ScanTalk demonstrates effectiveness in animating diverse 3D face meshes, including non-human variants. Such versatility holds significant promise for applications spanning video game development and virtual reality animation.

\subsection{User Study Interface}\label{sec:interface}
\begin{figure}[ht!]
    \centering
    \includegraphics[width=0.7\linewidth]{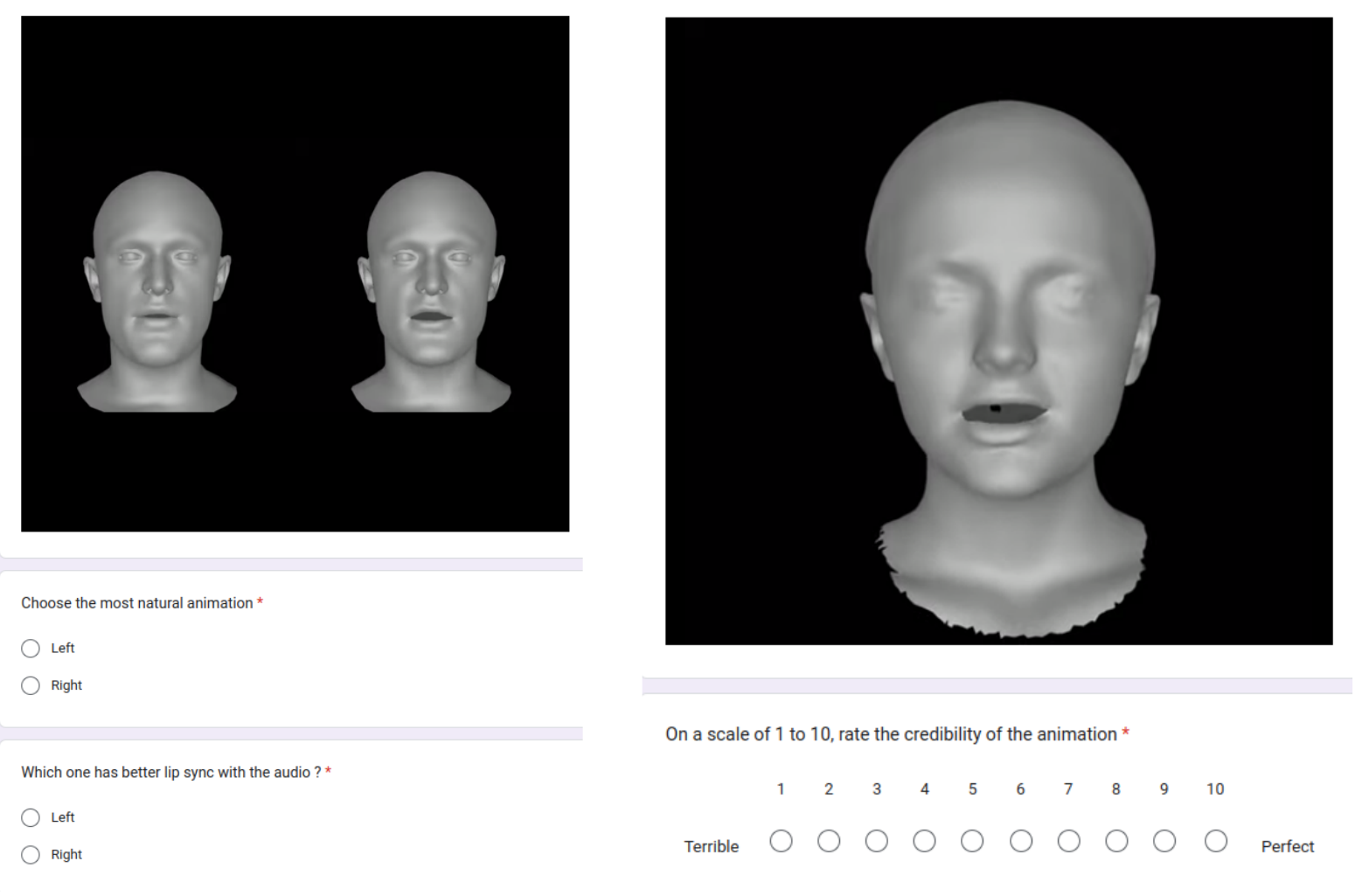}
    \caption{Examples of questions asked during the user study. (Left) Test 1, an A/B test to compare ScanTalk against state-of-the-art models. (Right) Test 2, we asked the users to evaluate the credibility of scan animations generated by ScanTalk.}   
    \label{fig:user_study_questions}
\end{figure}

In~\cref{fig:user_study_questions}, we depict the interface presented to users during our User Study detailed in Section 4.6 of the main paper. On the left, the interface for Test 1, an A/B test, is displayed, while the interface for Test 2 is showcased on the right.

\newpage
\bibliographystyle{plain}
\bibliography{bibliography.bib}

\begin{thebibliography}{10}

\bibitem{NeuralJacobianField_2022}
Noam Aigerman, Kunal Gupta, Vladimir~G. Kim, Siddhartha Chaudhuri, Jun Saito, and Thibault Groueix.
\newblock Neural jacobian fields: Learning intrinsic mappings of arbitrary meshes.
\newblock {\em ACM Trans. Graph.}, 41(4), jul 2022.

\bibitem{alghamdi2022_2Dtalking-head}
Mohammed~M. Alghamdi, He~Wang, Andrew~J. Bulpitt, and David~C. Hogg.
\newblock Talking head from speech audio using a pre-trained image generator.
\newblock In {\em Proceedings of the 30th ACM International Conference on Multimedia}, 2022.

\bibitem{Bahri_SMF_2021}
Mehdi Bahri, Eimear O’~Sullivan, Shunwang Gong, Feng Liu, Xiaoming Liu, Michael~M. Bronstein, and Stefanos Zafeiriou.
\newblock Shape my face: Registering 3d face scans by surface-to-surface translation.
\newblock {\em International Journal of Computer Vision (IJCV)}, Sep 2021.

\bibitem{varifold_loss}
Thomas Besnier, Sylvain Arguill{\`e}re, Emery Pierson, and Mohamed Daoudi.
\newblock {Toward Mesh-Invariant 3D Generative Deep Learning with Geometric Measures}.
\newblock {\em {Computers and Graphics}}, 2023.

\bibitem{Charles_PointNet_2017}
R.~Qi Charles, Hao Su, Mo~Kaichun, and Leonidas~J. Guibas.
\newblock Pointnet: Deep learning on point sets for 3d classification and segmentation.
\newblock In {\em 2017 IEEE Conference on Computer Vision and Pattern Recognition (CVPR)}, page 77–85. IEEE, Jul 2017.

\bibitem{Chen2020_2DtalkingheadGW}
Lele Chen, Guofeng Cui, Celong Liu, Zhong Li, Ziyi Kou, Yi~Xu, and Chenliang Xu.
\newblock Talking-head generation with rhythmic head motion.
\newblock In {\em European Conference on Computer Vision}, 2020.

\bibitem{Chen2021WavLMLS}
Sanyuan Chen, Chengyi Wang, Zhengyang Chen, Yu~Wu, Shujie Liu, Zhuo Chen, Jinyu Li, Naoyuki Kanda, Takuya Yoshioka, Xiong Xiao, Jian Wu, Long Zhou, Shuo Ren, Yanmin Qian, Yao Qian, Micheal Zeng, and Furu Wei.
\newblock Wavlm: Large-scale self-supervised pre-training for full stack speech processing.
\newblock {\em IEEE Journal of Selected Topics in Signal Processing}, 16:1505--1518, 2021.

\bibitem{Cosi_procedural_2002}
P.~Cosi, E.M. Caldognetto, G.~Perin, and C.~Zmarich.
\newblock Labial coarticulation modeling for realistic facial animation.
\newblock In {\em Proceedings. Fourth IEEE International Conference on Multimodal Interfaces}, pages 505--510, 2002.

\bibitem{Croquet_Diff_Reg_OT_2021}
Balder Croquet, Daan Christiaens, Seth~M. Weinberg, Michael Bronstein, Dirk Vandermeulen, and Peter Claes.
\newblock Unsupervised diffeomorphic surface registration and non-linear modelling.
\newblock In {\em Medical Image Computing and Computer Assisted Intervention (MICCAI)}, page 118–128. Springer, 2021.

\bibitem{VOCA2019}
Daniel Cudeiro, Timo Bolkart, Cassidy Laidlaw, Anurag Ranjan, and Michael Black.
\newblock Capture, learning, and synthesis of {3D} speaking styles.
\newblock In {\em Proceedings IEEE Conf. on Computer Vision and Pattern Recognition (CVPR)}, pages 10101--10111, 2019.

\bibitem{Dipanjan_2020_2Dtalkinghead}
Dipanjan Das, Sandika Biswas, Sanjana Sinha, and Brojeshwar Bhowmick.
\newblock Speech-driven facial animation using cascaded gans for learning of motion and texture.
\newblock In Andrea Vedaldi, Horst Bischof, Thomas Brox, and Jan-Michael Frahm, editors, {\em Computer Vision -- ECCV 2020}, pages 408--424, Cham, 2020. Springer International Publishing.

\bibitem{JALI_2016}
Pif Edwards, Chris Landreth, Eugene Fiume, and Karan Singh.
\newblock Jali: an animator-centric viseme model for expressive lip synchronization.
\newblock {\em ACM Trans. Graph.}, 35(4), jul 2016.

\bibitem{morphable_models_review}
Bernhard Egger, William A.~P. Smith, Ayush Tewari, Stefanie Wuhrer, Michael Zollhoefer, Thabo Beeler, Florian Bernard, Timo Bolkart, Adam Kortylewski, Sami Romdhani, Christian Theobalt, Volker Blanz, and Thomas Vetter.
\newblock {3D Morphable Face Models - Past, Present and Future}.
\newblock {\em {ACM Transactions on Graphics}}, 39(5):157:1--38, August 2020.

\bibitem{Fan_Lin_Saito_Wang_Komura_faceformer_2022}
Yingruo Fan, Zhaojiang Lin, Jun Saito, Wenping Wang, and Taku Komura.
\newblock Faceformer: Speech-driven 3d facial animation with transformers.
\newblock In {\em 2022 IEEE/CVF Conference on Computer Vision and Pattern Recognition (CVPR)}, page 18749–18758, New Orleans, LA, USA, Jun 2022. IEEE.

\bibitem{BIWI_2010}
Gabriele Fanelli, Juergen Gall, Harald Romsdorfer, Thibaut Weise, and Luc Van~Gool.
\newblock A 3-d audio-visual corpus of affective communication.
\newblock {\em IEEE Transactions on Multimedia}, 12(6):591--598, 2010.

\bibitem{gong2019spiralnet++}
Shunwang Gong, Lei Chen, Michael Bronstein, and Stefanos Zafeiriou.
\newblock Spiralnet++: A fast and highly efficient mesh convolution operator.
\newblock In {\em Proceedings of the IEEE International Conference on Computer Vision Workshops}, pages 0--0, 2019.

\bibitem{Texas3D_2010}
Shalini Gupta, Kenneth~R. Castleman, Mia~K. Markey, and Alan~C. Bovik.
\newblock Texas 3d face recognition database.
\newblock In {\em 2010 IEEE Southwest Symposium on Image Analysis \& Interpretation (SSIAI)}, pages 97--100, 2010.

\bibitem{facexhubert}
Kazi~Injamamul Haque and Zerrin Yumak.
\newblock Facexhubert: Text-less speech-driven e(x)pressive 3d facial animation synthesis using self-supervised speech representation learning.
\newblock In {\em INTERNATIONAL CONFERENCE ON MULTIMODAL INTERACTION (ICMI ’23)}, New York, NY, USA, 2023. ACM.

\bibitem{Hubert_audio_encoding}
Wei-Ning Hsu, Benjamin Bolte, Yao-Hung~Hubert Tsai, Kushal Lakhotia, Ruslan Salakhutdinov, and Abdelrahman Mohamed.
\newblock Hubert: Self-supervised speech representation learning by masked prediction of hidden units.
\newblock {\em IEEE/ACM Trans. Audio, Speech and Lang. Proc.}, 29:3451–3460, oct 2021.

\bibitem{EAMM_2022_2Dtalkinghead}
Xinya Ji, Hang Zhou, Kaisiyuan Wang, Qianyi Wu, Wayne Wu, Feng Xu, and Xun Cao.
\newblock Eamm: One-shot emotional talking face via audio-based emotion-aware motion model.
\newblock In {\em ACM SIGGRAPH 2022 Conference Proceedings}, SIGGRAPH '22, 2022.

\bibitem{adam}
Diederik~P. Kingma and Jimmy Ba.
\newblock Adam: A method for stochastic optimization, 2017.

\bibitem{Li_2020_Dynamic_facial_asset_and_rig_generation_from_a_single_scan}
Jiaman Li, Zhengfei Kuang, Yajie Zhao, Mingming He, Karl Bladin, and Hao Li.
\newblock Dynamic facial asset and rig generation from a single scan.
\newblock {\em ACM Trans. Graph.}, 39(6), nov 2020.

\bibitem{FLAME:SiggraphAsia2017}
Tianye Li, Timo Bolkart, Michael.~J. Black, Hao Li, and Javier Romero.
\newblock Learning a model of facial shape and expression from {4D} scans.
\newblock {\em ACM Transactions on Graphics, (Proc. SIGGRAPH Asia)}, 36(6):194:1--194:17, 2017.

\bibitem{spiralconv_Lim_2019}
Isaak Lim, Alexander Dielen, Marcel Campen, and Leif Kobbelt.
\newblock A simple approach to intrinsic correspondence learning on unstructured 3d meshes.
\newblock page 349–362, Berlin, Heidelberg, 2019. Springer-Verlag.

\bibitem{Liu_2019}
F.~Liu, L.~Tran, and X.~Liu.
\newblock 3d face modeling from diverse raw scan data.
\newblock In {\em 2019 IEEE/CVF International Conference on Computer Vision (ICCV)}, pages 9407--9417, Los Alamitos, CA, USA, nov 2019. IEEE Computer Society.

\bibitem{mediapipe_2019}
Camillo Lugaresi, Jiuqiang Tang, Hadon Nash, Chris McClanahan, Esha Uboweja, Michael Hays, Fan Zhang, Chuo-Ling Chang, Ming Yong, Juhyun Lee, Wan-Teh Chang, Wei Hua, Manfred Georg, and Matthias Grundmann.
\newblock Mediapipe: A framework for perceiving and processing reality.
\newblock In {\em Third Workshop on Computer Vision for AR/VR at IEEE Computer Vision and Pattern Recognition (CVPR) 2019}, 2019.

\bibitem{dominance_Massaro_procedural_2001}
Dom Massaro, Michael Cohen, Marija Tabain, Jonas Beskow, and R.~Clark.
\newblock Animated speech: Research progress and applications.
\newblock {\em Audiovisual Speech Processing}, 01 2001.

\bibitem{muralikrishnan_2023_BLISS}
Sanjeev Muralikrishnan, Chun-Hao~Paul Huang, Duygu Ceylan, and Niloy~J. Mitra.
\newblock Bliss: Bootstrapped linear shape space, 2023.

\bibitem{landmarks_3D_Nocentini_2023}
Federico Nocentini, Claudio Ferrari, and Stefano Berretti.
\newblock Learning landmarks motion from speech for speaker-agnostic 3d talking heads generation.
\newblock In Gian~Luca Foresti, Andrea Fusiello, and Edwin Hancock, editors, {\em Image Analysis and Processing -- ICIAP 2023}, pages 340--351, Cham, 2023. Springer Nature Switzerland.

\bibitem{nocentini2024emovocaspeechdrivenemotional3d}
Federico Nocentini, Claudio Ferrari, and Stefano Berretti.
\newblock Emovoca: Speech-driven emotional 3d talking heads, 2024.

\bibitem{peng2023selftalk}
Ziqiao Peng, Yihao Luo, Yue Shi, Hao Xu, Xiangyu Zhu, Hongyan Liu, Jun He, and Zhaoxin Fan.
\newblock Selftalk: A self-supervised commutative training diagram to comprehend 3d talking faces.
\newblock In {\em Proceedings of the 31st ACM International Conference on Multimedia}, page 5292–5301, 2023.

\bibitem{qi2017pointnet++}
Charles~Ruizhongtai Qi, Li~Yi, Hao Su, and Leonidas~J Guibas.
\newblock Pointnet++: Deep hierarchical feature learning on point sets in a metric space.
\newblock {\em Advances in neural information processing systems (NeurIPS)}, 30, 2017.

\bibitem{Qin_2023_NFR}
Dafei Qin, Jun Saito, Noam Aigerman, Groueix Thibault, and Taku Komura.
\newblock Neural face rigging for animating and retargeting facial meshes in the wild.
\newblock In {\em SIGGRAPH 2023 Conference Papers}, 2023.

\bibitem{COMA:ECCV18}
Anurag Ranjan, Timo Bolkart, Soubhik Sanyal, and Michael~J. Black.
\newblock Generating {3D} faces using convolutional mesh autoencoders.
\newblock In {\em European Conference on Computer Vision (ECCV)}, pages 725--741, 2018.

\bibitem{richard2021meshtalk}
Alexander Richard, Michael Zollh\"ofer, Yandong Wen, Fernando de~la Torre, and Yaser Sheikh.
\newblock Meshtalk: 3d face animation from speech using cross-modality disentanglement.
\newblock In {\em Proceedings of the IEEE/CVF International Conference on Computer Vision (ICCV)}, pages 1173--1182, October 2021.

\bibitem{bosphorus_2008}
Arman Savran, Ne{\c{s}}e Aly{\"u}z, Hamdi Dibeklio{\u{g}}lu, Oya {\c{C}}eliktutan, Berk G{\"o}kberk, B{\"u}lent Sankur, and Lale Akarun.
\newblock Bosphorus database for 3d face analysis.
\newblock In Ben Schouten, Niels~Christian Juul, Andrzej Drygajlo, and Massimo Tistarelli, editors, {\em Biometrics and Identity Management}, pages 47--56, Berlin, Heidelberg, 2008. Springer Berlin Heidelberg.

\bibitem{wav2vec_2019}
Steffen Schneider, Alexei Baevski, Ronan Collobert, and Michael Auli.
\newblock wav2vec: Unsupervised pre-training for speech recognition.
\newblock In {\em Interspeech 2019}, page 3465–3469. ISCA, September 2019.

\bibitem{sharp2021diffusion}
Nicholas Sharp, Souhaib Attaiki, Keenan Crane, and Maks Ovsjanikov.
\newblock Diffusionnet: Discretization agnostic learning on surfaces.
\newblock {\em ACM Trans. Graph.}, 01(1), 2022.

\bibitem{FaceDiffuser_Stan_MIG2023}
Stefan Stan, Kazi~Injamamul Haque, and Zerrin Yumak.
\newblock Facediffuser: Speech-driven 3d facial animation synthesis using diffusion.
\newblock In {\em ACM SIGGRAPH Conference on Motion, Interaction and Games (MIG '23), November 15--17, 2023, Rennes, France}, New York, NY, USA, 2023. ACM.

\bibitem{thambiraja2023_3diface}
Balamurugan Thambiraja, Sadegh Aliakbarian, Darren Cosker, and Justus Thies.
\newblock 3diface: Diffusion-based speech-driven 3d facial animation and editing, 2023.

\bibitem{Thambiraja_2023_ICCV_imitator}
Balamurugan Thambiraja, Ikhsanul Habibie, Sadegh Aliakbarian, Darren Cosker, Christian Theobalt, and Justus Thies.
\newblock Imitator: Personalized speech-driven 3d facial animation.
\newblock In {\em Proceedings of the IEEE/CVF International Conference on Computer Vision (ICCV)}, pages 20621--20631, October 2023.

\bibitem{Wang_2007_Rulebased_coarticulation_procedural}
Alice Wang, Michael Emmi, and Petros Faloutsos.
\newblock Assembling an expressive facial animation system.
\newblock In {\em Proceedings of the 2007 ACM SIGGRAPH Symposium on Video Games}, Sandbox '07, page 21–26, New York, NY, USA, 2007. Association for Computing Machinery.

\bibitem{Wang2023Emotional_2Dtalkinghead}
Jianrong Wang, Yaxin Zhao, Li~Liu, Tian-Shun Xu, Qi~Li, and Sen Li.
\newblock Emotional talking head generation based on memory-sharing and attention-augmented networks.
\newblock {\em ArXiv}, abs/2306.03594, 2023.

\bibitem{wang2021one_2Dtalkinghead}
Suzhen Wang, Lincheng Li, Yu~Ding, and Xin Yu.
\newblock One-shot talking face generation from single-speaker audio-visual correlation learning.
\newblock In {\em AAAI 2022}, 2022.

\bibitem{wuu2022multiface}
Cheng-hsin Wuu, Ningyuan Zheng, Scott Ardisson, Rohan Bali, Danielle Belko, Eric Brockmeyer, Lucas Evans, Timothy Godisart, Hyowon Ha, Xuhua Huang, Alexander Hypes, Taylor Koska, Steven Krenn, Stephen Lombardi, Xiaomin Luo, Kevyn McPhail, Laura Millerschoen, Michal Perdoch, Mark Pitts, Alexander Richard, Jason Saragih, Junko Saragih, Takaaki Shiratori, Tomas Simon, Matt Stewart, Autumn Trimble, Xinshuo Weng, David Whitewolf, Chenglei Wu, Shoou-I Yu, and Yaser Sheikh.
\newblock Multiface: A dataset for neural face rendering.
\newblock In {\em arXiv}, 2022.

\bibitem{xing2023codetalker}
Jinbo Xing, Menghan Xia, Yuechen Zhang, Xiaodong Cun, Jue Wang, and Tien-Tsin Wong.
\newblock Codetalker: Speech-driven 3d facial animation with discrete motion prior.
\newblock In {\em Proceedings of the IEEE/CVF Conference on Computer Vision and Pattern Recognition}, pages 12780--12790, 2023.

\bibitem{Xu_2013_diphone_coarticulation_procedural}
Yuyu Xu, Andrew~W. Feng, Stacy Marsella, and Ari Shapiro.
\newblock A practical and configurable lip sync method for games.
\newblock In {\em Proceedings of Motion on Games}, MIG '13, page 131–140, New York, NY, USA, 2013. Association for Computing Machinery.

\bibitem{yi2022_talkshow}
Hongwei Yi, Hualin Liang, Yifei Liu, Qiong Cao, Yandong Wen, Timo Bolkart, Dacheng Tao, and Michael~J Black.
\newblock Generating holistic 3d human motion from speech.
\newblock In {\em CVPR}, 2023.

\bibitem{BU3DFE_2006}
Lijun Yin, Xiaozhou Wei, Yi~Sun, Jun Wang, and M.J. Rosato.
\newblock A 3d facial expression database for facial behavior research.
\newblock In {\em 7th International Conference on Automatic Face and Gesture Recognition (FGR06)}, pages 211--216, 2006.

\bibitem{Zhong_2023_CVPR_2Dtalkinghead}
Weizhi Zhong, Chaowei Fang, Yinqi Cai, Pengxu Wei, Gangming Zhao, Liang Lin, and Guanbin Li.
\newblock Identity-preserving talking face generation with landmark and appearance priors.
\newblock In {\em Proceedings of the IEEE/CVF Conference on Computer Vision and Pattern Recognition (CVPR)}, pages 9729--9738, June 2023.

\end{thebibliography}

\end{document}